\pdfoutput=1
\newif\iffull
\fulltrue

\documentclass[11pt,twoside]{article}
\usepackage{xspace,fullpage,amsmath,amsthm,amssymb,amsfonts,boxedminipage,microtype,hyperref}
\usepackage{paralist}
\usepackage{bm}
\usepackage{bbm}
\usepackage{array}
\usepackage{multirow}
\usepackage{times}
\usepackage{graphicx}
\usepackage{mathtools}
\usepackage{float}
\usepackage{subcaption}
\usepackage[ruled,vlined]{algorithm2e}
\DontPrintSemicolon

\usepackage{color}
\usepackage[style=alphabetic,backend=bibtex,maxbibnames=40,maxcitenames=6,giveninits=true,doi=false,url=true]{biblatex}
\newcommand*{\citet}[1]{\AtNextCite{\AtEachCitekey{\defcounter{maxnames}{2}}} \textcite{#1}}

\newcommand*{\citep}[1]{\cite{#1}}

\bibliography{ra-bib}
\makeatletter

\providecommand{\R}{{\mathbb R}}

\providecommand{\E}{\mathop{\mathbf E}}

\providecommand{\ti} \tilde

\providecommand{\eps}{\epsilon}

\providecommand{\ignore}[1]{\relax}

\providecommand{\eat}[1]{}
\providecommand{\remove}[1]{}

\newtheorem{thm}{Theorem}[section]
\newtheorem{lem}[thm]{Lemma}
\newtheorem{clm}[thm]{Claim}
\newtheorem{cor}[thm]{Corollary}
\newtheorem{rem}[thm]{Remark}

\newtheorem{defn}[thm]{Definition}

\makeatother
\makeatletter
\newcommand{\out}{y}

\newcommand{\outDom}{\mathcal{Y}}
\newcommand{\alg}{M}

\newcommand{\domain}{\mathcal{X}}
\newcommand{\dataset}{S}
\newcommand{\view}{\boldsymbol{v}}
\newcommand{\viewcount}{\boldsymbol{vc}}

\newcommand{\lvec}[1]{\reflectbox{$\vec{\reflectbox{$#1$}}$}}

\renewcommand{\eps}{\varepsilon}

\newcommand{\loss}{\ell}
\newcommand{\lossFunc}[3]{\loss\left(#1; #2, #3 \right)}

\newcommand{\lossRV}{L}
\newcommand{\lossRVfunc}[2]{\lossRV_{#1, #2}}

\newcommand{\HockeyStick}[1]{\boldsymbol{H}_{#1}}

\newcommand{\HockeyStickDiv}[3]{\HockeyStick{#1}\left(#2 ~\left\Vert~ #3 \right. \right)}

\newcommand{\Pois}[2]{\mathcal{P}_{#1}\left(#2\right)}

\newcommand{\alloc}[2]{\mathcal{A}_{#1}\left(#2\right)}
\newcommand{\allocFunc}[3]{\mathcal{A}_{#1}\left(#2; #3\right)}

\newcommand{\PoisGen}[3]{\mathcal{P}_{#1}^{#2}\left(#3\right)}
\newcommand{\allocGen}[3]{\mathcal{A}_{#1}^{#2}\left(#3\right)}

\newcommand{\PoisMask}[3]{\mathcal{PM}_{#1}^{#2}\left(#3\right)}
\newcommand{\allocMask}[3]{\mathcal{AM}_{#1}^{#2}\left(#3\right)}

\newcommand{\NNI}{\mathbb{Z}_+}
\newcommand{\add}[1]{\lvec{#1}}
\newcommand{\remv}[1]{\vec{#1}}

\newcommand{\privProf}[1]{\delta_{#1}}
\newcommand{\privProfAdd}[1]{\add{\delta}_{#1}}
\newcommand{\privProfRem}[1]{\remv{\delta}_{#1}}

\newcommand{\Bin}{\mathrm{Bin}}

\newcommand{\meanopt}{{\mu^{\ast}}}
\newcommand{\sigmastar}{{\sigma^{\ast}}}
\newcommand{\logdelta}{\lambda}

\newtheorem{prop}[thm]{Proposition}

\newcommand{\simRem}{\mathrel{\substack{\sim\\[-0.8ex] \scalebox{0.6}{$\rightarrow$}}}}
\makeatother

\begin{document}

\title{Counting on Dummies: Privacy Amplification with Observable Counts and Truncated Batches}
\author{
Vitaly Feldman \\Apple \and Audra McMillan \\ Apple \and Kunal Talwar \\ Apple
}
\maketitle

\begin{abstract}

    In many practical deployments of Differentially Private Stochastic Gradient Descent (DP-SGD), the server processes data in batches. Randomly sampling these batches allows for better privacy bounds via privacy amplification results. Most sampling schemes with provable privacy amplification, including Poisson sampling and random allocation, result in variable batch sizes. The privacy guarantee of these schemes requires that the batch sizes are not revealed. In distributed settings, the server performing the aggregation or any party observing the communication traffic may learn these batch sizes. Furthermore, in both centralized and distributed settings, memory and computation constraints may require each batch to be of some maximum size $s$, necessitating truncation of the batch when it exceeds this limit. It is known that this truncation can result in a significant increase in privacy loss. In this work, we address both of these challenges via a simple mechanism that uses a random number of \emph{dummy points} to privatize the batch size and maintain the privacy guarantee even under truncation. We demonstrate that our mechanism is easy to implement in a distributed setting, and provide a general analysis of this mechanism for both Poisson sampling and random allocation. Our analysis is designed to support efficient and relatively tight privacy accounting. It also effectively amplifies the privacy of the count releases by subsampling, making the expected number of dummy points asymptotically smaller than what is needed to directly privatize the (sensitivity-one) batch sizes with non-negative noise. For Poisson sampling with truncated batches, we demonstrate empirically that our mechanism achieves better privacy--efficiency tradeoffs than the direct accounting for truncation recently given by \citet{Ganesh25}.

\end{abstract}

\section{Introduction}

Differentially private stochastic gradient descent (DP-SGD) is the standard algorithm for training machine-learning models under differential privacy \citep{ACGMMT16,BST14}. Its privacy analysis rests on \emph{privacy amplification by subsampling}: because each step operates on a small randomly chosen {\em batch}, the relatively weak privacy guarantee of a single noisy gradient step is amplified significantly.  This enables the release of the numerous noisy gradients with reasonable privacy guarantees for the entire run. The strongest and most widely used amplification comes from schemes in which each batch is a genuinely random subset of the data: \emph{Poisson subsampling}, which includes each example independently with probability $\gamma$ \citep{ACGMMT16,ZDW22}, and, more recently, \emph{random allocation}, a balls-in-bins scheme that spreads each example across a random subset of the steps \citep{FS25ra}. Both schemes admit tight and efficient accounting, based on dominating pairs of distributions \citep{ZDW22} and implemented via numerical computations on privacy loss distributions (PLDs).%

An important feature of these schemes is that they produce batches of \emph{variable size}, distributed as a binomial random variable. The size of a batch is thus a random quantity that depends on the internal randomization of the algorithm. The amplification guarantees of these schemes implicitly assume that the batch size is \emph{never revealed}, that is, that the adversary observes only the per-step outputs. Under a trusted central curator this assumption is typically harmless. However, in federated and distributed deployments, where contributions from many devices are aggregated without a trusted central party, the server that performs the aggregation typically \emph{does} observe each batch's size, for example, by counting the number of messages it combines. Once the size is observed, the amplification analysis no longer applies as stated, and the resulting leakage erodes the privacy guarantees. Related gaps between the idealized sampling assumed in analyses and the batching actually realized in implementations are by now a recognized source of privacy violations \citep{CGKKMSZ24a,ABDCH24,LRKS24}. Indeed, in federated settings the difficulty of realizing and hiding subsampling has led to the use of optimization methods that abandon sampling altogether \citep{KMSTTX21}.

Variable batch sizes present another, independent difficulty. Memory and compute budgets typically impose an \emph{a priori} cap $s$ on the batch size, so any batch exceeding $s$ must be \emph{truncated} to fit. In common training pipeline designs, the batch also needs to be padded to have the fixed size $s$. Similarly in distributed aggregation implementations, system constraints can force a maximum batch size. Truncation, however, breaks the privacy accounting under the add/remove and zero-out adjacency notions that are typically used in this context. When the batch size is fixed (or otherwise known), removing a data point from a full batch means that another data point takes its place, so the batch sum can shift by two gradients rather than one. This effectively doubles the sensitivity and thus requires doubling the noise scale. Recent work of \citet{Ganesh25} gives a tighter, dedicated analysis of truncated Poisson sampling, but the doubled sensitivity remains: it is charged whenever the batch overflows, which forces the expected batch size to be well below $s$. We provide a more detailed discussion in Section~\ref{sec:related}.

\subsection{Our results}
We resolve both these difficulties with a single, simple technique that we call \emph{dummy count masking}. Throughout, the \emph{count} of a batch is the number of contributions it holds: without truncation this is exactly the batch size, and with truncation it is the pre-truncation size. Before releasing the count, we add a random number of \emph{dummy points} to the batch. A dummy is a fixed reference element $\bot$, which in aggregation is simply the zero vector; for this reason, replacing a real point by $\bot$ is referred to as \emph{zero-out} adjacency, and the privacy of such a replacement is controlled by the core mechanism. Crucially, the number of dummy points added is drawn from a fixed, public, data-independent distribution $C$ over $\NNI$. We then release the resulting (dummy-masked) count. In addition, if a truncation cap is imposed, we truncate the resulting batch to the fixed size $s$.  If a fixed size $s$ is required we also pad the batches with fewer than $s$ elements using the same dummy element.

This has two effects. First, privatizing the count reduces to releasing a sensitivity-one integer statistic with one-sided noise, a problem studied in several contexts \citep{CHM21,XWSD25}. Second, conditioned on the released count, a removed user is replaced by an additional dummy rather than by another user's data, so the two neighboring batches differ in a single entry even after truncation and padding. Truncation thus preserves unit sensitivity instead of creating a sensitivity-two event.

Importantly, dummy count masking is compatible with distributed aggregation without a trusted curator. Each user decides independently, by local coin flips, whether to inject extra dummy messages into a batch and how many. With anonymous aggregation, dummy reports are indistinguishable from real ones, so the total number of dummies is hidden. Under distributed noise generation, each dummy message can carry its own noise share, so the scheme composes naturally with client-side noise generation and integrates easily into existing secure-aggregation pipelines.

We give a general privacy analysis of this mechanism for both Poisson subsampling and random allocation, showing that the additional privacy cost is typically dominated by the cost of the original algorithm. Our guarantees are stated as explicit dominating pairs of distributions, so that accurate numerical privacy accounting can be applied directly. We describe the simple steps needed to compute the privacy loss distribution of our mechanism and then demonstrate empirically that its privacy cost is typically negligible. We also compare the efficiency of our mechanism empirically with that of truncation without masking \citep{Ganesh25}.

\subsection{Overview of the analysis}
We show (Section~\ref{sec:batch}) that both the release of the masked count and the truncation can be folded into the per-step algorithm: the dummy-masked and (optionally) truncated scheme is \emph{exactly} an ordinary sampling scheme applied to a modified \emph{batch algorithm}. We then give a \emph{dominating pair} of distributions for this batch algorithm in terms of a dominating pair for the original per-step mechanism, where a dominating pair of a mechanism is a pair of distributions whose hockey-stick divergence upper bounds the privacy loss of the mechanism. Specifically, if $(P_0,Q_0)$ is a remove-direction dominating pair of the per-step mechanism (the Gaussian noisy sum being the canonical example), we show that $(C_{+1}\times P_0,\,C\times Q_0)$ is a dominating pair of the modified batch algorithm, where $C$ is the distribution of the number of dummies and $C_{+1}$ is the distribution of $R+1$ for $R\sim C$. As a result, we never need to re-derive amplification theorems: it suffices to feed this dominating pair to the existing subsampling \citep{ZDW22} and random-allocation \citep{FS25ra} results.

Truncation is handled by isolating a stability property (Section~\ref{sec:batch-trunc}). Conditioned on the released count, padding or truncating to size $s$ is a data-independent mixture of \emph{stable} maps, that is, maps that transform batches differing in a single entry to batches differing in a single entry. By joint convexity of the hockey-stick divergence \citep{BBG18}, pre-composing with such a map can only \emph{decrease} every divergence, so it preserves dominating pairs (Lemma~\ref{lem:adjpres}). The dominating pair for the batch algorithm above is therefore unchanged by truncation (Lemma~\ref{lem:batchpair}).

\paragraph{Participant-generated dummies.} In distributed applications of our mechanism the requirement that dummies are generated independently of batch participation may create a significant overhead in that each client may need to be actively communicating dummies across the entire execution of the algorithm. In particular, they may need to contribute dummies during rounds in which they are not contributing data. Thus we also analyze the version in which only the users that join a batch inject dummies into it (Section~\ref{sec:participant}). In this case the count noise is coupled to participation. We show that our analysis framework can also handle this case (albeit with a different dominating pair of distributions). The analysis relies on an additional domination monotonicity argument for the batch size and a minimum-batch-participation event charged to~$\delta$.

 \paragraph{The cost of masking.} The main computational overhead of our mechanism is the expected number of dummies per batch, $\E_{R\sim C}[R]$. Released on its own, a count with \emph{one-sided} (non-negative) noise requires noise of mean $\Theta(\tfrac1\eps\log\tfrac{\eps}\delta)$ \citep{CHM21,XWSD25}. Combining this with black-box amplification and composition theorems gives a per-batch mean of $O(\sigmastar \log(\gamma t/\delta))$ for $\sigmastar:=\gamma\sqrt{2t\log(1/\delta)}/\eps$, where $\gamma$ is the Poisson subsampling rate and $t$ is the number of steps. In Section~\ref{sec:onesided} we show that the $\log(1/\delta)$ factor can be avoided in the CLT regime, where the privacy of the subsampled and composed count release is tightly controlled by the $\chi^2$ divergence of a single step \citep{FS26clt}. We then show that, under the simplifying assumptions of the CLT regime, when masking the counts of a subsampled and composed or randomly allocated Gaussian mechanism with noise multiplier $\sigma$,  a mean of $O\bigl(\tfrac{\sigma}{\sqrt\alpha}\log\tfrac{\sigma}{\sqrt\alpha}\bigr)$ dummies suffices to ensure that the final $\eps$ grows by at most a factor of $1+\alpha$.

\paragraph{Why not simply privatize the counts?} For one epoch of random allocation, the vector of batch counts is itself a sensitivity-one statistic, since deleting a user decrements a single count by one. One might therefore hope to privatize it on its own, by adding noise to each count, and simply pay for this release via standard composition theorems. The subtlety is that the two parts of the release share their randomness. The allocation that determines the counts is precisely the randomness whose secrecy drives privacy amplification for the noisy gradient sums, so once the (even noisy) counts are public, the amplification guarantee for the sums no longer applies as stated. More generally, it is well known that two correlated releases that are each differentially private may jointly reveal the input completely.\footnote{As a simple example, let $\mathcal{A}_1(D)=f(D)+N$ for some function $f$ and noise $N\sim\nu$ with a suitably chosen distribution $\nu$, and let $\mathcal{A}_2(D)=N$. Each of $\mathcal{A}_1$ and $\mathcal{A}_2$ is DP on its own, but together they reveal $f(D)$ exactly.}

By folding the count release into the per-batch algorithm, we analyze the two components jointly, conditioning both on the shared sampling randomness. The joint analysis also gives better bounds: it effectively applies privacy amplification by subsampling to the count releases themselves, since a user's presence affects any given count only with probability about $1/t$. By the asymptotics of Section~\ref{sec:onesided}, the amplified count release needs noise of mean $\tilde O\left(\frac{1}{\sqrt{t}\eps}\right)$ per batch, smaller by a factor of $\tilde\Omega(\sqrt{t})$ than the $\tilde \Theta(\tfrac1\eps)$ required by an unamplified sensitivity-one release with non-negative noise.

 \paragraph{Why non-negative noise only?} Since the primary goal of masking is to privatize the counts, it is natural to ask why we do not allow negative count noise: whenever the count noise $R$ is negative, one could drop $|R|$ randomly chosen elements from the batch. This is implementable when the server can manipulate batches arbitrarily before computing on them. It does not, however, fit our main motivation, the observability of message counts in a distributed setting with user-generated dummies, since a user cannot drop messages generated by other users. Moreover, our analysis conditions on the released count. With non-negative noise, the two conditional batches can be coupled as zero-out neighbors, so that the divergence of the outputs is controlled by the core mechanism. With deletions, conditioning on the count would change which real elements remain in the batch, and this coupling would no longer be possible.

\paragraph{Numerical Evaluation.} By combining our results with numerical PLD accounting methods in  \citep{Google20,FS26fft}, we evaluate the privacy overheads of count masking empirically in two ways. We first consider two parameter settings corresponding to multi-epoch DP-SGD with both Poisson subsampling and random allocation. We consider several count noise distributions, including the simple binomial and negative binomial distributions (which can be generated in a distributed setting) with optimized parameters, and the count noise distribution optimized for minimizing the second moment under a fixed mean from \citep{XWSD25}. In all settings, with the optimized negative binomial, the privacy overhead of masking is negligible ($<1\%$) once the expected number of dummies equals the standard deviation of the batch size. The overhead of the simple binomial is higher but still modest. We also evaluate the setting in which dummies are generated only by the participants of a batch: there, the overhead is essentially identical to that of independent dummies for larger batches and moderately higher for smaller ones (Figure~\ref{fig:exp-participant}).

We then consider the truncation setting where we have fixed maximum batch size. We compare the privacy cost of plain truncation via the analysis of \citet{Ganesh25} with that of truncation after dummy masking, both under Poisson sampling. We show that empirically, adding dummies achieves a significantly lower privacy cost at the same batch efficiency (the expected fraction of real elements in the truncated batch). Importantly, count masking also makes truncation possible under random allocation, for which no alternative truncation analysis exists. Details of the experiments are provided in Section~\ref{sec:experiments}.

\subsection{Related work}
\label{sec:related}
\paragraph{Batch sampling in theory and practice.}
The gap between the sampling that DP-SGD analyses assume and the batching that implementations realize is by now well documented: batching by shuffling can result in substantially weaker privacy guarantees than the Poisson subsampling that is usually used for accounting \citep{CGKKMSZ24a,CGKKMSZ24b}, and similar gaps arise for other deployed batching choices \citep{ABDCH24,LRKS24}. Fixed-size pipelines have led practitioners to implement Poisson subsampling with truncation and padding \citep{CGKKMSZ24b}, the route taken in recent large-scale differentially private language-model training \citep{MHSBCCGKKL25,VaultGemma25} and analyzed tightly by \citet{Ganesh25}. Random allocation  was proposed and analyzed as an alternative with more predictable participation \citep{CGHLKKMSZ24,CCGHST25, DCO25, FS25ra}, with efficient PLD accounting given by \citet{FS26pld}. All of these analyses assume that the batch counts are hidden.

\paragraph{Truncated Poisson sampling.}
The closest prior work on privacy amplification with batch padding and truncation is that of \citet{Ganesh25}, who gives a tight privacy bound for \emph{truncated Poisson sampling} of the subsampled Gaussian mechanism: each example is included independently with probability $p$, and a batch exceeding the cap $B$ is replaced by a uniformly random size-$B$ subset before the noisy gradient sum is released. Like ours, the analysis is phrased through dominating pairs, and it improves substantially on charging the overflow event to $\delta$, which would force the expected batch size well below $B$. Its central observation is that truncation \emph{doubles the sensitivity}: when an added example displaces another from an already-full batch, the released sum can move by two gradients rather than one. The resulting dominating pair is a public mixture of an untruncated sensitivity-$1$ component and a truncated sensitivity-$2$ component, weighted by the (data-independent) overflow probability, so the doubling is paid only when truncation can occur. Our approach is orthogonal, and in principle the two can be combined; Section~\ref{sec:experiments} compares them empirically. Our analysis also applies beyond the Gaussian mechanism, to any per-step algorithm with a remove-direction dominating pair, and covers random allocation alongside Poisson subsampling.

\paragraph{Dummy messages in differential privacy.}
Injecting randomly generated \emph{dummy} messages  is a recurring idea in distributed and shuffle-model protocols. \citet{BBGN20,GKMP20} realize private summation and counting in the shuffle model by having users send extra messages whose random contents implement the privacy noise, and study which noise distributions can be generated this way at the smallest message cost. \citet{LLFHLRQ21} amplify the privacy of shuffle-model protocols with dummy points, and in the augmented shuffle protocols of \citet{MSE25} the shuffler itself injects, per item, a binomial (or asymmetric two-sided geometric) number of dummies so that the dummies alone realize the estimator's noise. Dummies also defend against traffic analysis: Vuvuzela's mix servers add random cover traffic so that the observable dead-drop access counts are differentially private \citep{Vuvuzela15}, and in the private-search system Wally each client issues fake queries over an anonymous network so that per-shard query counts reveal little \citep{Wally24}.

Several works in this context also study {\em infinitely divisible} count-noise laws, which can be generated in a distributed way by summing independent contributions from many users. In particular, our use of negative-binomial distributions is inspired by \citet{GKMP20}.

\paragraph{One-sided noise.}
Optimal non-negative noise for a single sensitivity-one release is well understood: \citet{CHM21} analyze simple families such as the truncated geometric, negative binomial and Poisson padding. \citet{XWSD25} show that the $\Theta(\tfrac1\eps\log\tfrac1\delta)$ mean error achieved by a clamped two-sided geometric distribution is the minimum possible. For repeated releases \citet{XWSD25} develop a hybrid R\'enyi analysis and a convex program for composition-optimal one-sided noise.
Our experiments include this noise and show that the negative binomial, which can be generated in a distributed setting, achieves comparable privacy guarantees.

\section{Definitions}\label{sec:defs}

We write $[t]\coloneqq\{1,\ldots,t\}$ and $\NNI\coloneqq\{0,1,2,\ldots\}$.

We describe privacy guarantees in terms of the privacy loss random variable \citep{DR16} and the closely related hockey-stick privacy profile \citep{BBG18}.
    Given two distributions $P, Q$, the \emph{privacy loss random variable} $\lossRVfunc{P}{Q}$ is defined by $\lossFunc{\omega}{P}{Q} \coloneqq \ln \left(\frac{P(\omega)}{Q(\omega)}\right)$ where $\omega \sim P$. We refer to its distribution as the \emph{privacy loss distribution (PLD)}.
    For $\kappa\geq 0$, the $\kappa$-\emph{hockey-stick divergence} between two distributions $P, Q$ is defined as
    \[
        \HockeyStickDiv{\kappa}{P}{Q} \coloneqq \E_{\omega \sim P}\left[\bigl[1 - \kappa \cdot e^{-\lossFunc{\omega}{P}{Q}} \bigr]_{+}\right],
    \]
    where $\left[x \right]_{+} \coloneqq \max\{0, x\}$ \cite{BKOZB12}.

    The dataset adjacency notions we consider are the standard add/remove and zero-out notions. In add/remove adjacency datasets $\dataset, \dataset' \in \domain^{*}$ are adjacent if $\dataset$ can be obtained from $\dataset'$ via adding or removing a single element. For the zero-out notion, datasets $\dataset, \dataset' \in \domain^{*}$ are adjacent if $\dataset$ can be obtained from $\dataset'$ by replacing a single element with the ``zero'' element $\bot$ which we assume to be in $\domain$.  In the common application of aggregation this element is just $0$ (or the zero vector).

    Given an algorithm $\alg : \domain^{*} \to \outDom$, its privacy profile $\privProf{\alg} : \R \to [0, 1]$ is defined as
    $\privProf{\alg}(\eps)\coloneqq\sup_{\dataset,\dataset'}\HockeyStickDiv{e^{\eps}}{\alg(\dataset)}{\alg(\dataset')}$, where the supremum is over all pairs of adjacent datasets.
    Since the hockey-stick divergence is asymmetric in general, we write $\privProfRem{\alg}$ for the supremum restricted to the \emph{remove} direction, in which $\dataset'$ is obtained from $\dataset$ by removing an element or replacing it with $\bot$, and $\privProfAdd{\alg}$ for the \emph{add} direction, in which $\dataset$ is obtained from $\dataset'$ in this way. Consequently, $\privProf{\alg}(\eps) = \max\{\privProfRem{\alg}(\eps), \privProfAdd{\alg}(\eps) \}$.  Given $\eps > 0$, $\delta \in [0, 1 ]$, an algorithm $\alg$ will be called \emph{$(\eps, \delta)$-differentially private (DP)} if $\privProf{\alg}(\eps) \le \delta$ \citep{DKMMN06}.

    A key concept for characterizing the privacy guarantees of an algorithm is that of a \emph{dominating pair} of distributions \citep{ZDW22}. Given distributions $P_{1}, Q_{1}, P_{2}, Q_{2}$, we say that a pair of distributions $(P_1,Q_1)$ \emph{dominates} the pair of distributions $(P_{2},Q_{2})$ if for all $\kappa \ge 0$ we have $\HockeyStickDiv{\kappa}{P_{2}}{Q_{2}} \le \HockeyStickDiv{\kappa}{P_{1}}{Q_{1}}$. If $\privProfRem{\alg}(\eps) \le \HockeyStickDiv{e^{\eps}}{P}{Q}$ for every $\eps \in \R$, we call $(P, Q)$ a \emph{remove-direction dominating pair} for $\alg$. The swapped pair controls the add direction: $\privProfAdd{\alg}(\eps) \le \HockeyStickDiv{e^{\eps}}{Q}{P}$ for every $\eps \in \R$ \citep{ZDW22}. This ``swap'' fact, which we use repeatedly, relies on domination at all $\kappa\ge0$: it follows from the identity $\HockeyStickDiv{\kappa}{Q}{P}=\kappa\,\HockeyStickDiv{1/\kappa}{P}{Q}+1-\kappa$, valid for all distributions $P,Q$ and all $\kappa>0$.%

      Dominating pairs compose adaptively \citep{ZDW22}: if $(P_1,Q_1)$ dominates $M$ and, for every output $o$ of $M$, $(P_2,Q_2)$ dominates $M'(\cdot,o)$, then $(P_{1} \times P_{2}, Q_{1}\times Q_{2})$ dominates the adaptive composition that releases $o=M(\dataset)$ followed by $M'(\dataset,o)$. For independent draws, the privacy loss of the product pair is the sum of the component privacy losses in distribution: $\lossRVfunc{P_{1} \times P_{2}}{Q_{1} \times Q_{2}} \stackrel{d}{=} \lossRVfunc{P_{1}}{Q_{1}} + \lossRVfunc{P_{2}}{Q_{2}}$.

\subsection{Sampling schemes}

    We consider \emph{$t$-step algorithms} built from a per-step algorithm $\alg$ that takes a subset of the dataset and the sequence of previous outputs as its input. We denote the domain of its input \emph{elements} by $\domain$ and the set of its possible \emph{outputs} by $\outDom$. Formally, denoting $\outDom^{<t} = \bigcup_{i<t} \outDom^{i}$, $\alg$ takes a dataset in $\domain^{*}$ and a view $\view \in\outDom^{<t}$ as its inputs, and outputs a value in $\outDom$. A $t$-step algorithm using $\alg$ first uses some scheme to define $t$ subsets $\dataset^{1}, \ldots, \dataset^{t} \subseteq \dataset$, then sequentially computes $\out_{i} = \alg \left(\dataset^{i}, \view_{1:i-1} \right)$, where $\view_{1:i} \coloneqq (\out_{1}, \ldots, \out_{i})$ are the intermediate \emph{views} consisting of the outputs produced so far, and $\view_{1:0} = \emptyset$.

    The elements of $\dataset$ are assigned to the subsets by a \emph{sampling scheme}. We consider two common schemes: Poisson subsampling and random allocation.%

\paragraph{Poisson subsampling}
The \emph{Poisson scheme} is parameterized by a sampling parameter $\gamma\in[0,1]$. Each element of the dataset is included in each subset $S_i$ with probability $\gamma$ independently of other elements and subsets.
For a $t$-step algorithm defined by an algorithm $\alg$, we denote by $\Pois{t, \gamma}{\alg} : \domain^{*} \to \outDom^{t}$ the resulting algorithm when $\alg$ is used with the Poisson subsampling scheme.

 A dominating pair for a single step of Poisson subsampling is obtained by mixing a dominating pair of the original algorithm.
 \begin{lem}[\citep{ZDW22}]\label{lem:subsamPLDdom}
        Given $\gamma \in [0,1]$ and an algorithm $\alg$, denote by $\alg \circ \mathcal{P}$ the algorithm that, given a dataset $\dataset$, constructs a subset $\dataset'$ by independently including each element in $\dataset$ w.p. $\gamma$ then releases $\alg(\dataset')$. If $\alg$ is dominated by the pair of distributions $(P, Q)$ in the remove direction, then $\alg \circ \mathcal{P}$ is dominated by $(P_{\gamma}, Q)$ in the remove direction and by $(Q, P_{\gamma})$ in the add direction, where $P_{\gamma} \coloneqq \gamma P + (1-\gamma) Q$.
\end{lem}
 Since the $t$ steps of Poisson subsampling form an adaptive composition, a dominating pair for the $t$-step algorithm is the product of $t$ copies of the single-step pair. We can therefore focus on a single step, $t=1$, which we denote simply by $\Pois{\gamma}{\alg}$.

\paragraph{Random allocation}
The random allocation scheme of \citet{FS25ra}, parameterized by a number of selected steps $k\in[t]$, assigns each element $x_l$ to a uniformly random $k$-subset $\bm{i}_l\subseteq[t]$ (independently across elements, with all $\binom{t}{k}$ subsets equally likely). We then define the subsets $\dataset^{i}:=\{x_l : i\in\bm{i}_l\}$ and the outputs $\out_i=\alg(\dataset^i,\view_{1:i-1})$ are produced sequentially as above. We denote by $\alloc{t,k}{\alg}:\domain^{*}\to\outDom^{t}$ the resulting $t$-step algorithm and omit $k$ from the notation when $k=1$. Because step $i$ receives the view $\view_{1:i-1}$, the scheme is an adaptive composition of its $t$ steps; see \cite[Section~2.1]{FS25ra}.

\begin{defn}[Random allocation of a pair]\label{def:RApair}
For distributions $P,Q$ on a common space $\outDom$ and integers $1\le k\le t$, the \emph{$k$-out-of-$t$ random allocation} of $(P,Q)$, written $\mathsf{RA}_{t,k}(P,Q)$, is the distribution on $\outDom^{\,t}$ obtained by drawing a uniformly random $k$-subset $\bm{i}\subseteq[t]$ and sampling the coordinates independently, the $j$-th from $P$ if $j\in\bm{i}$ and from $Q$ otherwise. We write $Q^{\otimes t}$ for the $t$-fold product of $Q$, the distribution on $\outDom^{\,t}$ with every coordinate drawn independently from $Q$.
\end{defn}

The privacy of the allocation scheme is governed by any per-step remove-direction dominating pair, through the following result \cite[Theorem~3.1]{FS25ra}.

\begin{lem}[\citep{FS25ra}]\label{lem:RAdom}
If the per-step algorithm $\alg$ is dominated by $(P,Q)$ in the remove direction for every view, then $(\mathsf{RA}_{t,k}(P,Q),\,Q^{\otimes t})$ is a dominating pair for $\alloc{t,k}{\alg}$ under add/remove adjacency. In particular, for all add/remove neighbors $S\supsetneq S'$ with $|S|=|S'|+1$ and every $\kappa\ge0$,
\[
    \HockeyStickDiv{\kappa}{\allocFunc{t,k}{\alg}{S}}{\allocFunc{t,k}{\alg}{S'}}
    \ \le\
    \HockeyStickDiv{\kappa}{\mathsf{RA}_{t,k}(P,Q)}{Q^{\otimes t}},
\]
where $\mathsf{RA}_{t,k}(P,Q)$ and $Q^{\otimes t}$ are as in Definition~\ref{def:RApair}. Since $(\mathsf{RA}_{t,k}(P,Q),Q^{\otimes t})$ is a dominating pair, the add direction is controlled by the swapped pair, $\HockeyStickDiv{\kappa}{\allocFunc{t,k}{\alg}{S'}}{\allocFunc{t,k}{\alg}{S}}\le\HockeyStickDiv{\kappa}{Q^{\otimes t}}{\mathsf{RA}_{t,k}(P,Q)}$ for all $\kappa\ge0$.
\end{lem}

\begin{rem}
    Dominating pairs for subsampling and random allocation are often stated using zero-out adjacency for both $\alg$ and the sampling scheme (e.g.~\cite{FS25ra}). The zero-out versions assume that including $\bot$ in a batch does not affect the output distribution of $\alg$; under this assumption add/remove and zero-out adjacency are equivalent, and both versions of Lemmas~\ref{lem:subsamPLDdom} and \ref{lem:RAdom} hold. The batch algorithms we define, however, release the (masked) count of the batch and are therefore sensitive to the inclusion of $\bot$. We thus rely explicitly on add/remove adjacency when applying privacy amplification, while assuming privacy under zero-out adjacency for the core batch algorithm, since dummy count masking adds $\bot$ elements to the batch.
\end{rem}

\section{Setup}\label{sec:setup}

Given a dataset $S=(x_1,\ldots,x_n)$ and a $t$-step algorithm defined by a per-step algorithm $\alg$ (Section~\ref{sec:defs}), we study two modifications of the sampling process: releasing the count, and fixing the batch size by padding or truncation. We assume that $\domain$ contains a ``zero'' element $\bot$, which defines the zero-out adjacency notion and is used for padding. We do not assume that $\bot$ has no effect on $\alg$; in particular, we allow algorithms such as mean estimation whose output depends on the total number of elements, $\bot$'s included.

\paragraph{Truncated sampling:}
We first define the padded/truncated versions of the sampling schemes. In these schemes $\alg$ operates only on batches of a fixed size $s$. Given a batch $U$ of size $c$, the scheme pads it with $\bot$ elements to size $s$ if $c<s$, and truncates it to a uniformly random subset of $s$ elements if $c>s$, before feeding the result to $\alg$. We define these versions as $\PoisGen{t,\gamma}{=s}{\alg}$ for the Poisson subsampling case, and $\allocGen{t,k}{=s}{\alg}$ for the random allocation case. We refer to these sampling schemes  as {\em $s$-truncated}.

\paragraph{Public count:}
We say that a sampling scheme has \emph{public count} if, in addition to the outputs of $\alg$, it releases at each step the number of elements in the batch. Without truncation this count is exactly the size of the batch that $\alg$ processes; with truncation $\alg$ processes a batch of size $s$ while the released count $c_i$ is the pre-truncation size. We also allow $\alg$ to use the counts: we assume that $c_1,\ldots,c_i$ are part of the view $\view$ given to $\alg$ at step $i$, and emphasize this by writing the view as $\viewcount \in \outDom^{<t} \times\NNI^{\leq t}$.

\paragraph{Dummy count masking:}
Our main tool for controlling the privacy cost of these two modifications is adding a random number of dummy elements to each batch. Given a count distribution $C$ over $\NNI$, the \emph{$C$-masked} version of a sampling scheme draws, at each step $i$, an independent $r_i\sim C$ and adds $r_i$ dummy elements $\bot$ to the sampled batch $\dataset^i$ before feeding it to $\alg$. This masking can be applied to both the Poisson and random allocation sampling schemes and we denote these as $\PoisMask{t,\gamma}{C}{\alg}$ and $\allocMask{t,k}{C}{\alg}$ respectively. The public-count version of a masked scheme releases the count after the dummies are added, and the $s$-truncated version truncates or pads the batch after the dummies are added. We denote the $C$-masked, $s$-truncated Poisson and random allocation schemes by $\PoisMask{t,\gamma}{C,=s}{\alg}$ and $\allocMask{t,k}{C,=s}{\alg}$, respectively.

\subsection{Distributed implementation for aggregation}\label{sec:distributed}

We specialize the schemes of this section to private \emph{high-dimensional aggregation}, the primitive underlying DP-SGD, federated sum, and mean estimation. There are $n$ clients; client $i$ holds a vector $x_i\in\R^d$, and at each of $t$ steps $j$ we wish to release a private estimate of the sum $\sum_{i\in \dataset^j}x_i$ of the vectors in the sampled batch $\dataset^j$. The per-step primitive is \emph{noisy summation}: the batch sum perturbed by additive noise whose magnitude is set by the target privacy level. We keep the noise distribution and any norm bound on the $x_i$ general, abstracting the per-step mechanism by its remove-direction dominating pair $(P_0,Q_0)$ (Section~\ref{sec:defs}). With $\alg$ being the noisy sum, the sampling, truncation and dummy count masking of this section yield a $t$-step mechanism whose output for each batch is the masked (pre-truncation) count $c$ together with the noisy sum $y$ of the processed batch.%

In the central model, a trusted curator would, at each step, form the batch by the sampling scheme, mask its count by inserting a $C$-distributed number of dummy entries, truncate to size $s$, and release $(c,y)$. We obtain the \emph{same} released distribution without a trusted curator, in two parts. First, the sampling is itself distributed. Each client use its local randomness to decide which batches to participate in, with independent per-step choices for Poisson subsampling, or the random $k$-subset for random allocation. Second, both the privacy noise and the dummy entries can be generated by the clients, with each client adding a share of the Gaussian noise.  We emphasize that our information-theoretic guarantees are for the released per-batch aggregates $(c,y)$ and a secure aggregation protocol and an anonymization layer are needed (typically under cryptographic assumptions) for this model to be appropriate. Such secure aggregation protocols have been studied under a variety of attack models; see, e.g., \cite{DKMMN06, Bonawitz17,Corrigan-GibbsB17, Talwaretal24,JinCMRYO24,eichner2024confidential}. We describe the sampling, noise and dummy generation mechanisms below and then give concrete examples of client algorithms.

\subsubsection{Distributed noise generation (fixed batch size)}\label{sec:noise}
The released sum of a batch is privatized by additive noise. Each contribution to the batch, real or dummy, is sent with an independent noise share already added, so user $i$ sends $x_i$ plus its share of the noise. This setting is easiest to analyze when the total noise in a batch does not depend on the number of contributions $c$ received by the server. This is achieved by truncation when $c\ge s$ together with padding when $c<s$: the protocol must then generate $s-c$ padding dummies carrying independent noise shares, or add the missing noise shares within the secure aggregation.

\subsubsection{Distributed dummy generation}\label{sec:dummy}
For each batch, each user independently draws $d_i\sim\rho$ and sends $d_i$ dummy contributions to it, whether or not it joins the batch with its own data. The per-batch dummy count is then $C=\rho^{*n}$, the distribution of $\sum_{i\in[n]} d_i$.

Dummy generation is independent of the data and, for now, of the participation decisions. The privacy analysis uses only the resulting abstraction, which is the same for central and distributed dummies and for both sampling schemes: each batch independently receives $R\sim C$ dummy ($\bot$) contributions, for a fixed, public distribution $C$ over $\NNI$.

\paragraph{Adjacency for distributed dummies.}
With central dummies, $C$ is fixed by the server and the guarantees hold under add/remove adjacency as stated. With distributed dummies, $C=\rho^{*n}$ is generated by the $n$ participating devices, so the full departure of a user from the protocol would also remove one $\rho$-draw from every batch. The privacy guarantees therefore correspond to the deletion of a user's \emph{data} while its device continues to run the (data-independent) dummy protocol. Accounting for the full departure of a device from the protocol would additionally charge, for every batch, the shift of the count by one $\rho$-draw, which is not amplified by sampling. We leave this extension to future work.

\subsubsection{Example: Gaussian client code with Bernoulli dummies}
We instantiate the client side for the \emph{Gaussian} case with \emph{distributed} noise and \emph{distributed} dummies. Here $\|x_i\|\le 1$ and the noise is spherical: each contribution, real or dummy, carries an independent $N(\bar 0,\sigma_1^2 I_d)$ share, so the sum of a size-$s$ batch carries noise $N(\bar 0,s\sigma_1^2 I_d)$, and a dummy is a pure-noise message. Dummy counts are binomial: with a per-user bias $q\in[0,1]$, every user sends a dummy to each batch independently with probability $q$, so $\rho=\mathrm{Bern}(q)$ and the per-batch dummy count is $C=\rho^{*n}=\mathrm{Binomial}(n,q)$. Algorithms~\ref{alg:client-poisson} and~\ref{alg:client-ra} describe the algorithm run by client $i$ under Poisson subsampling and under random allocation, respectively.

\begin{algorithm}[H]
\caption{Client $i$: Poisson subsampling (Gaussian, distributed noise, binomial dummies)}\label{alg:client-poisson}
\KwIn{point $x_i\in\R^d$; sampling rate $\gamma$; dummy bias $q$; noise scale $\sigma_1$; steps $t$}
\For{$j\gets 1$ \KwTo $t$}{
  \If{$\mathrm{Bern}(\gamma)=1$}{
    draw $\zeta\sim N(\bar 0,\sigma_1^2 I_d)$ and send $x_i+\zeta$ to batch $j$ \tcp*{data}
  }
  \If{$\mathrm{Bern}(q)=1$}{
    draw $\zeta'\sim N(\bar 0,\sigma_1^2 I_d)$ and send $\zeta'$ to batch $j$ \tcp*{dummy}
  }
}
\end{algorithm}

\begin{algorithm}[H]
\caption{Client $i$: Random allocation (Gaussian, distributed noise, binomial dummies)}\label{alg:client-ra}
\KwIn{point $x_i\in\R^d$; participation count $k$; dummy bias $q$; noise scale $\sigma_1$; steps $t$}
Sample $A_i\subseteq[t]$ uniformly at random with $|A_i|=k$\;
\ForEach{$j\in A_i$}{
  draw $\zeta_j\sim N(\bar 0,\sigma_1^2 I_d)$ and send $x_i+\zeta_j$ to batch $j$ \tcp*{data}
}
\For{$j\gets 1$ \KwTo $t$}{
  \If{$\mathrm{Bern}(q)=1$}{
    draw $\zeta'_j\sim N(\bar 0,\sigma_1^2 I_d)$ and send $\zeta'_j$ to batch $j$ \tcp*{dummy}
  }
}
\end{algorithm}

The data and dummy decisions are independent, so in either scheme each batch receives $\mathrm{Binomial}(n,q)$ dummies regardless of how many real points it draws.%

\subsubsection{Server side}\label{sec:server}
Each batch is processed separately. The server receives the set of (real and dummy) contributions in a uniformly random order. It keeps the first $s$ and pads with dummies if fewer than $s$ arrive, and releases the count $c$ together with their sum $y$.
Each of the $t$ steps yields one batch, so the released transcript is the collection $(c_1,y_1),\dots,(c_t,y_t)$.%

\section{The batch algorithm and its dominating pair}\label{sec:batch}

The key insight behind our analysis is that both modifications of the sampling schemes can be viewed as modifications of the algorithm $\alg$ acting on a single batch. We can therefore derive a dominating pair for the new batch algorithm and then obtain the privacy of each scheme in Section~\ref{sec:amp} from the corresponding amplification lemma. The dummies act as noise added to the count released by the batch algorithm, making the count differentially private. Once the masked count is released, truncation is a data-independent transformation of the batch that preserves adjacency; it therefore preserves any dominating pair of the pre-truncation algorithm and incurs no additional privacy loss.

We treat two variants in turn: the \emph{public-count} batch algorithm $M_C$, which releases the dummy-masked count and runs $\alg$ on the whole batch, and its \emph{$s$-truncated} variant $M_{C,s}$, which additionally truncates the batch to a fixed size $s$ before running $\alg$.  We do not analyze the private-count, truncated-only variant separately; its release without the count is a postprocessing of $M_{C,s}$. Both variants have the same dominating pair.%

\paragraph{The public-count batch algorithm $M_C$.}
Fix a count distribution $C$ over $\NNI$ and a per-step algorithm $\alg$. The \emph{public-count} batch algorithm $M_C=M_C[\alg]$ is the per-batch core of the $C$-masked, public-count schemes of Section~\ref{sec:setup}. It takes the sequence $U$ of points assigned to the batch and, in the adaptive setting, a view $v$ of the previously released batch outputs, and produces an output $(c,o)$:
\begin{enumerate}
\item draw a dummy count $R\sim C$; the batch $B$ consists of the points of $U$ together with $R$ dummy elements (each equal to $\bot$), and its count $c:=|U|+R$ is released (the public count);
\item release $o:=\alg(B,(v,c))$, the output of $\alg$ run on the whole batch $B$ of these $c$ elements, with the count $c$ available to $\alg$ as part of its view.
\end{enumerate}
A dummy is a copy of $\bot$, counted in $c$ and handed to $\alg$ as one of the batch's elements. The dummy count $R\sim C$ is internal to $M_C$, drawn afresh and independently in each invocation, and does not depend on $U$ or $v$. We analyze $M_C$ under add/remove (deletion) adjacency: $U_-$ is $U$ with one point deleted, so $|U_-|=|U|-1$ and the released count $c$ is one smaller.

\paragraph{The $s$-truncated algorithm $M_{C,s}$.}
The \emph{$s$-truncated} batch algorithm $M_{C,s}=M_{C,s}[\alg]$ is the per-batch core of the $C$-masked, $s$-truncated, public-count schemes of Section~\ref{sec:setup}.
It is similar to $M_C$ except that it inserts a truncation step before $\alg$. After drawing $R\sim C$ and releasing the count $c$, it truncates or pads the batch $B$ to a batch $\hat{B}$ of size exactly $s$. If $c\ge s$, this is achieved by keeping a uniformly random $s$-subset of $B$, and if $c<s$ this is achieved by padding with $(s-c)$ additional dummies. As in $M_C$, it releases $o:=\alg(\hat{B},(v,c))$. Because $\alg$ now sees only $s$ elements, it need only be defined, and dominated, on inputs of the fixed size $s$.

The following lemma states that the masked public-count schemes with $\alg$ are exactly the basic sampling schemes with the batch algorithms defined above.

\begin{lem}[Reduction to the batch algorithm]\label{lem:reduction}
Fix a per-step algorithm $\alg$ and a count distribution $C$, and let $M_C=M_C[\alg]$ and $M_{C,s}=M_{C,s}[\alg]$ be the public-count and $s$-truncated batch algorithms defined above (each releases its count). Then, on every dataset and as distributions over transcripts, the $C$-masked public-count schemes of Section~\ref{sec:setup} are exactly the basic sampling schemes run on the batch algorithm,
\[
\PoisMask{t,\gamma}{C}{\alg}=\Pois{t,\gamma}{M_C},\qquad \allocMask{t,k}{C}{\alg}=\alloc{t,k}{M_C},
\]
and likewise their $s$-truncated versions $\PoisMask{t,\gamma}{C,=s}{\alg}$ and $\allocMask{t,k}{C,=s}{\alg}$ equal $\Pois{t,\gamma}{M_{C,s}}$ and $\alloc{t,k}{M_{C,s}}$.
\end{lem}
\begin{proof}
Both sides assign points to the $t$ batches by the same sampling scheme and then process each batch identically. The claim follows.
\end{proof}

By Lemma~\ref{lem:reduction} the privacy analysis of every scheme we consider reduces to understanding a remove-direction dominating pair for the batch algorithm, and then applying the appropriate privacy amplification analysis.

\subsection{Dominating pair: public count}\label{sec:batch-kb}
We now exhibit a remove-direction dominating pair for $M_C$. It is built from a remove-direction dominating pair $(P_0,Q_0)$ of $\alg$ under zero-out adjacency. Explicitly, we assume that \emph{for every view $\viewcount$ that includes the released counts} and all inputs $W,W_\bot$ that agree except in one entry, which holds some element of $\domain$ in $W$ and $\bot$ in $W_\bot$, the pair $(\alg(W,\viewcount),\alg(W_\bot,\viewcount))$ is dominated by $(P_0,Q_0)$. When $\bot$ does not affect the output of $\alg$, this is the same as a remove-direction dominating pair under add/remove adjacency.

\begin{lem}[Dominating pair, public count]\label{lem:batchpair-kb}
Suppose the per-step algorithm $\alg$ is dominated by $(P_0,Q_0)$ in the remove direction of zero-out adjacency, for every view that includes the released count. Set
\[
    P:=C_{+1}\times P_0,\qquad Q:=C\times Q_0,
\]
where $C_{+1}$ is the distribution of $R+1$ for $R\sim C$. Then for every view $v$ and every pair of add/remove neighbors $U,U_-$, with $U_-$ obtained from $U$ by deleting one element $x_n$, $(M_C(U,v),M_C(U_-,v))$ is dominated by $(P,Q)$ and $(M_C(U_-,v),M_C(U,v))$ by $(Q,P)$.
\end{lem}
\begin{proof}
Fix $v$ and write $M_C(\cdot)=M_C(\cdot,v)$; its output is the pair $(c,o)$. We exhibit $M_C$ as the adaptive composition of two stages and dominate each. The first stage is releasing the count $c$, the second is outputting $o$ given $c$. Let $a:=|U_-|$, so $U$ has $a+1$ points and $U_-$ has $a$; write $K_1(W)$ for the distribution of $c$ and $K_2(W,c)$ for the conditional distribution of $o$ given $c$.

\emph{Count.} With $R\sim C$, $K_1(U)$ and $K_1(U_-)$ are the distributions of $R+a+1$ and $R+a$. Since these are the translations by $a$ of $C_{+1}$ (the distribution of $R+1$) and $C$ (the distribution of $R$), and translation is a bijection that preserves every hockey-stick divergence, the count stage is dominated by $(C_{+1},C)$.%

\emph{Output given count.} Fix a count $c$. If $c<a+1$ it is unattainable on $U$ (i.e.\ $K_1(U)$ puts no mass at $c$); redefining $K_2(U,c):=K_2(U_-,c)$ there makes the pair identical and does not change the distribution of $M_C(U)$. So assume $c\ge a+1$, so that both batches have exactly $c$ elements. Align the two batches: both contain the $a$ common real points and $c-a-1$ common dummies, and the remaining entry is $x_n$ under $U$ and a dummy $\bot$ under $U_-$. The two size-$c$ batches $B,B_\bot$ fed to $\alg$ thus differ in a single entry. As $\alg$ receives the same view $(v,c)$ in both worlds, the hypothesis applied to that view-with-count gives $(K_2(U,c),K_2(U_-,c))=(\alg(B,(v,c)),\alg(B_\bot,(v,c)))$ dominated by $(P_0,Q_0)$, for every $c$.

The claim then follows by adaptive composition of dominating pairs \citep{ZDW22}.
\end{proof}

\subsection{Dominating pair: truncated batch size}\label{sec:batch-trunc}

The truncation step of $M_{C,s}$ is an instance of a simple operation: a data-independent random transformation of the batch that can only bring neighboring batches closer together. We isolate this operation and show that it preserves dominating pairs; a dominating pair for $M_{C,s}$ then follows by composing the count stage of Lemma~\ref{lem:batchpair-kb} with the transformed output stage. Throughout, an \emph{adjacency notion} on datasets is an arbitrary directed binary relation, written $W\simRem W'$, with the direction matching the remove direction: for example, $W\simRem W'$ when the two agree except in one entry, which holds some element of $\domain$ in $W$ and $\bot$ in $W'$.

\begin{defn}[Adjacency-preserving random map]\label{def:adjpres}
Let $T$ be a randomized map that takes an input dataset $W$ and coins $r\sim\theta$, where $\theta$ is a fixed distribution that does not depend on $W$. We say that $T$ \emph{preserves adjacency} if for every $r$ and every $W\simRem W'$, either $T(W,r)=T(W',r)$ or $T(W,r)\simRem T(W',r)$. Here $\simRem$ denotes an adjacency notion on the inputs of $T$ and one on its outputs; the two need not be the same relation (in our application the inputs have size $c$ and the outputs size $s$).
\end{defn}

Joint convexity of the hockey-stick divergence \citep{BBG18} then gives the following lemma.
\begin{lem}[Preprocessing by an adjacency-preserving map preserves domination]\label{lem:adjpres}
Let $A$ be an algorithm such that for every pair $V\simRem V'$ the pair $(A(V),A(V'))$ is dominated by $(P_0,Q_0)$, and let $T$, with coins $r\sim\theta$ drawn independently of $A$'s internal randomness, preserve adjacency. Denote by $A\circ T$ the algorithm that on input $W$ draws $r\sim\theta$ and releases $A(T(W,r))$. Then for every $W\simRem W'$ the pair $\bigl((A\circ T)(W),(A\circ T)(W')\bigr)$ is dominated by $(P_0,Q_0)$.%
\end{lem}

We now observe that dummy addition followed by truncation/padding is an adjacency-preserving map. Fix the target size $s\ge1$ and a batch size $c\ge1$, and view a size-$c$ batch as a tuple in $\domain^c$ (with $\bot\in\domain$); let size-$c$ adjacency $B\simRem B_\bot$ hold when the two tuples agree in all coordinates except one, which holds some $x\in\domain$ in $B$ and $\bot$ in $B_\bot$.

\begin{clm}[Truncation/padding preserves adjacency]\label{clm:truncadj}
Define $T_{c,s}(B,r)$ on $\domain^c$ as follows: if $c\ge s$, the coins $r$ select a uniformly random $s$-subset $I\subseteq[c]$ and $T_{c,s}$ keeps the coordinates in $I$; if $c<s$, $T_{c,s}$ deterministically appends $s-c$ dummies ($\bot$). Then $T_{c,s}$ preserves adjacency from size-$c$ neighbors to size-$s$ neighbors: for every $r$ and every $B\simRem B_\bot$, either $T_{c,s}(B,r)=T_{c,s}(B_\bot,r)$ or $T_{c,s}(B,r)\simRem T_{c,s}(B_\bot,r)$.
\end{clm}
\begin{proof}
The same coins $r$ act on the same coordinate positions in both worlds. If $c\ge s$ and the distinguished coordinate is not in $I$, the images are equal. On the other hand, when the distinguished coordinate is kept, or when $c<s$ and the appended dummies are common, the images agree in all coordinates except the distinguished one, which is $x$ in the image of $B$ and $\bot$ in the image of $B_\bot$. Thus the outcomes are size-$s$ neighbors, in the same direction.
\end{proof}

\begin{lem}[Dominating pair, truncated batch size]\label{lem:batchpair}
Suppose the per-step algorithm $\alg$ is dominated by $(P_0,Q_0)$ in the remove direction of zero-out adjacency on inputs of size $s$, for every view that includes the released count. Then, with $P:=C_{+1}\times P_0$ and $Q:=C\times Q_0$, for every view $v$ and every pair of add/remove neighbors $U,U_-$, with $U_-$ obtained from $U$ by deleting one element $x_n$, $(M_{C,s}(U,v),M_{C,s}(U_-,v))$ is dominated by $(P,Q)$ and $(M_{C,s}(U_-,v),M_{C,s}(U,v))$ by $(Q,P)$.
\end{lem}
\begin{proof}
With $a:=|U_-|$ as before, the count stage is dominated by $(C_{+1},C)$ exactly as in Lemma~\ref{lem:batchpair-kb}; only the output stage changes, because of truncation. Fix a count $c\ge a+1$ (the case $c<a+1$ is handled as before) and write $K_2(\cdot,c)$ for the conditional distribution of $o$ given $c$. Conditioned on the released count $c$, the pre-truncation batches are deterministic: aligning coordinates, $B$ consists of the $a$ common points, $c-a-1$ common dummies, and $x_n$ in the distinguished coordinate, while $B_\bot$ is identical with a dummy ($\bot$) in place of $x_n$.%
The output stage is exactly $A\circ T_{c,s}$ for $A:=\alg(\cdot,(v,c))$: the batch is truncated or padded to size $s$ by $T_{c,s}$, whose coins are independent of the data and of $\alg$'s internal randomness, and $\alg$ is then run on the result. By hypothesis $A$ is dominated by $(P_0,Q_0)$ on size-$s$ inputs differing in a single entry for every view $(v,c)$, and Claim~\ref{clm:truncadj} shows that $T_{c,s}$ preserves adjacency. Lemma~\ref{lem:adjpres} therefore establishes domination of $(K_2(U,c),K_2(U_-,c))$ by $(P_0,Q_0)$, for every $c$. Composing the two stages as in Lemma~\ref{lem:batchpair-kb} dominates $(M_{C,s}(U),M_{C,s}(U_-))$ by $(P,Q)$, and the swap fact of Section~\ref{sec:defs} gives the add direction $(Q,P)$.
\end{proof}

\begin{rem}\label{rem:adjpres-variants}
Lemma~\ref{lem:adjpres} isolates the only property of truncation the analysis uses, so Lemma~\ref{lem:batchpair} holds verbatim for any data-independent batch transformation that preserves adjacency: e.g.\ a different (data-independent) truncation rule, random reordering of the batch, or sub-sampling within the batch. Rules that on $c<s$ bypass $\alg$ altogether and release a data-independent default value are also covered, directly rather than through the lemma: on that event the two conditional distributions are identical, so the conditional divergence is $[1-\kappa]_+$, at most $\HockeyStickDiv{\kappa}{P_0}{Q_0}$. Taking $T$ to be the identity recovers the output stage of the public-count variant (Lemma~\ref{lem:batchpair-kb}).
\end{rem}

\paragraph{Gaussian instantiation.}
Take $\alg$ to be the Gaussian sum: on a batch of $s$ elements $w_1,\dots,w_s$ of $\ell_2$ norm at most $1$, where a dummy or padding element is the zero vector $\bar 0=\bot$, it releases $\sum_{\ell\le s} w_\ell+\xi$ with $\xi\sim N(\bar 0,s\sigma_1^2 I_d)$. The variance scales with $s$ because, under the distributed noise generation of Section~\ref{sec:noise}, each of the $s$ summands carries an independent $N(\bar 0,\sigma_1^2 I_d)$ share; equivalently, the server adds $N(\bar 0,s\sigma_1^2 I_d)$ centrally. Replacing one batch element by $\bot=\bar 0$ shifts the released mean by a vector of norm at most $1$ while leaving the noise $N(\bar 0,s\sigma_1^2 I_d)$ unchanged, so $\alg$ is dominated in the remove direction of zero-out adjacency by $(P_0,Q_0)=(N(1,s\sigma_1^2),N(0,s\sigma_1^2))$ \citep{ACGMMT16}. Lemma~\ref{lem:batchpair} then gives the batch dominating pair
\[
    P=C_{+1}\times N(1,s\sigma_1^2),\qquad Q=C\times N(0,s\sigma_1^2).
\]
Truncation is exactly what makes this pair independent of the batch size: $\alg$ always sums $s$ elements, whereas in the public-count variant $M_C$ the noise would be $N(\bar 0,c\sigma_1^2 I_d)$ and the per-step pair $(N(1,c\sigma_1^2),N(0,c\sigma_1^2))$ would depend on the public count $c$. This is not directly compatible with privacy amplification results that require a fixed dominating pair for the batch algorithm. However, accounting can still be supported via a monotonicity argument for domination with a lower bound on the batch size similar to the one we describe in Section~\ref{sec:participant}.

\section{Privacy amplification}\label{sec:amp}

By Lemma~\ref{lem:reduction}, each scheme is the basic sampling scheme applied to the batch algorithm of Section~\ref{sec:batch}, either $M_C$ or its truncated variant $M_{C,s}$, and by Lemmas~\ref{lem:batchpair-kb} and~\ref{lem:batchpair} both have the same remove-direction dominating pair $(P,Q)=(C_{+1}\times P_0,C\times Q_0)$. Amplification depends only on $(P,Q)$, so it applies verbatim to either variant and we state the results for the truncated version for concreteness. For a dataset $S=(x_1,\dots,x_n)$ and its deletion neighbor $S_-=(x_1,\dots,x_{n-1})$, we denote by $D(S)$ the output of the scheme under analysis: a single batch in the Poisson case and the $t$-tuple of batch outputs in the random allocation case. Each subsection amplifies $(P,Q)$ via the corresponding result of Section~\ref{sec:defs}, obtaining a dominating pair that controls both the remove and the add direction.

\subsection{Poisson subsampling}
The scheme is $\Pois{t,\gamma}{M_{C,s}}=\PoisMask{t,\gamma}{C,=s}{\alg}$, basic Poisson subsampling at rate $\gamma$ of the batch algorithm. It samples and dummy draws are independent across steps and the steps compose adaptively, so it suffices to analyze a single batch $M_{C,s}\circ\mathcal{P}_\gamma$ (Section~\ref{sec:defs}), whose dominating pair follows by applying Lemma~\ref{lem:subsamPLDdom} to the batch pair $(P,Q)$.

\begin{cor}[Poisson subsampling of the batch algorithm]\label{cor:poissonamp}
Fix a batch size $s$ and a per-batch count distribution $C$, and suppose the per-step algorithm $\alg$ satisfies the hypothesis of Lemma~\ref{lem:batchpair}: it is dominated by $(P_0,Q_0)$ in the remove direction of zero-out adjacency on size-$s$ inputs, for every view that includes the released counts. Let $D=M_{C,s}\circ\mathcal{P}_\gamma$ be a single step of the Poisson scheme $\PoisMask{t,\gamma}{C,=s}{\alg}=\Pois{t,\gamma}{M_{C,s}}$ at rate $\gamma$, where $M_{C,s}=M_{C,s}[\alg]$ is the batch algorithm of Section~\ref{sec:batch}. Then $(P_\gamma,Q)$ is a dominating pair for $D$ under add/remove adjacency, where
\[
    P:=C_{+1}\times P_0,\qquad Q:=C\times Q_0,\qquad P_\gamma:=\gamma P+(1-\gamma)Q,
\]
and $C_{+1}$ denotes the distribution of $R+1$ for $R\sim C$. That is, for any pair of deletion neighboring datasets $S$ and $S_-$, $(D(S),D(S_-))$ is dominated by $(P_\gamma,Q)$ and $(D(S_-),D(S))$ by $(Q,P_\gamma)$.
\end{cor}
\begin{proof}
By Lemma~\ref{lem:batchpair} the batch algorithm $M_{C,s}$ is dominated by $(P,Q)$ in the remove direction for every view. Applying the subsampling domination of Lemma~\ref{lem:subsamPLDdom} makes $(P_\gamma,Q)$ a dominating pair for $D=M_{C,s}\circ\mathcal{P}_\gamma$, so $(D(S),D(S_-))$ is dominated by $(P_\gamma,Q)$ and, by the swapped pair, $(D(S_-),D(S))$ by $(Q,P_\gamma)$.
\end{proof}

Consequently, the full adaptive $t$-step Poisson transcript is dominated (in the remove direction) by $(P_\gamma^{\otimes t},Q^{\otimes t})$.

\subsection{Random allocation}
Here the mechanism is $\alloc{t,k}{M_{C,s}}=\allocMask{t,k}{C,=s}{\alg}$, the random allocation scheme of Section~\ref{sec:defs} applied to the batch algorithm; its dominating pair follows by applying the allocation domination of \citet{FS25ra} (Lemma~\ref{lem:RAdom}) to the batch pair $(P,Q)$.

\begin{cor}[Random allocation of the batch algorithm]\label{cor:RAmain}
Fix a batch size $s$ and a per-batch count distribution $C$, and suppose the per-step algorithm $\alg$ satisfies the hypothesis of Lemma~\ref{lem:batchpair}: it is dominated by $(P_0,Q_0)$ in the remove direction of zero-out adjacency on size-$s$ inputs, for every view that includes the released counts. Let $D=\alloc{t,k}{M_{C,s}}=\allocMask{t,k}{C,=s}{\alg}$ be random allocation applied to the batch algorithm $M_{C,s}=M_{C,s}[\alg]$ of Section~\ref{sec:batch}. Then $(\mathsf{RA}_{t,k}(P,Q),\,Q^{\otimes t})$ is a dominating pair for $D$ under add/remove adjacency, where
\[
    P:=C_{+1}\times P_0,\qquad Q:=C\times Q_0,
\]
$C_{+1}$ denotes the distribution of $R+1$ for $R\sim C$, and $Q^{\otimes t}$ is the $t$-fold product of $Q$ (Definition~\ref{def:RApair}). That is, for any deletion neighboring datasets $S$ and $S_-$, $(D(S),D(S_-))$ is dominated by $(\mathsf{RA}_{t,k}(P,Q),Q^{\otimes t})$ and $(D(S_-),D(S))$ by $(Q^{\otimes t},\mathsf{RA}_{t,k}(P,Q))$.
\end{cor}
\begin{proof}
By Lemma~\ref{lem:batchpair} the batch algorithm $M_{C,s}$ is dominated by $(P,Q)$ in the remove direction for every view. Lemma~\ref{lem:RAdom} then makes $(\mathsf{RA}_{t,k}(P,Q),Q^{\otimes t})$ a dominating pair for $D=\alloc{t,k}{M_{C,s}}$, so $(D(S),D(S_-))$ is dominated by $(\mathsf{RA}_{t,k}(P,Q),Q^{\otimes t})$ and, by the swapped pair, $(D(S_-),D(S))$ by $(Q^{\otimes t},\mathsf{RA}_{t,k}(P,Q))$.
\end{proof}

Since the pair $(D(S),D(S_-))$ is dominated by the random-allocation pair $(\mathsf{RA}_{t,k}(P,Q),Q^{\otimes t})$, every accounting tool for random allocation applies directly. The remove-direction divergence $\HockeyStickDiv{\kappa}{\mathsf{RA}_{t,k}(P,Q)}{Q^{\otimes t}}$ can be efficiently evaluated numerically \citep{FS26pld,FS26fft}, and admits a closed-form bound \citep{FS25ra,FS26clt}. Up to lower-order terms it is at most the privacy profile of $t$-step Poisson subsampling of the batch pair $(P,Q)$ at rate $\eta\approx k/t$.
\section{Participant-generated dummies}\label{sec:participant}

In the distributed-dummy scheme of Section~\ref{sec:dummy} every user injects a $\rho$-distributed number of dummies into \emph{every} batch, whether or not it participates, so each batch carries an independent, data-independent count noise $C=\rho^{*n}$. In the distributed setting,
this is wasteful: a user sends data to only $k$ of the $t$ batches, yet must still send remain online and send dummy traffic to all $t$. We analyze the cheaper variant in which \emph{only the users that join a batch inject dummies into it}: each participant of a batch adds an independent $\rho$-distributed number of dummies to \emph{that} batch, so a user contributes to only the $k$ batches it joins, for both data and dummies.
We now analyze this case, where the count noise is coupled to the allocation.

Algorithm~\ref{alg:client-ra-participant} is the participant-generated analogue of Algorithm~\ref{alg:client-ra} (Gaussian noise and Bernoulli dummies under random allocation).
The server side is unchanged (Section~\ref{sec:server}).

\begin{algorithm}[H]
\caption{Client $i$: Random allocation with participant-generated dummies (Gaussian, distributed noise, Bernoulli dummies)}\label{alg:client-ra-participant}
\KwIn{point $x_i\in\R^d$; participation count $k$; dummy bias $q$; noise scale $\sigma_1$; steps $t$}
Sample $A_i\subseteq[t]$ uniformly at random with $|A_i|=k$\;
\ForEach{$j\in A_i$}{
  draw $\zeta_j\sim N(\bar 0,\sigma_1^2 I_d)$ and send $x_i+\zeta_j$ to batch $j$ \tcp*{data}
  \If{$\mathrm{Bern}(q)=1$}{
    draw $\zeta'_j\sim N(\bar 0,\sigma_1^2 I_d)$ and send $\zeta'_j$ to batch $j$ \tcp*{dummy}
  }
}
\end{algorithm}

As in Section~\ref{sec:batch}, the core mechanism $\alg$ receives the released count as part of its view.  We assume throughout this section that $\alg$ has a single remove-direction dominating pair $(P_0,Q_0)$ of zero-out adjacency.%

\paragraph{Participant-generated versions of the sampling schemes.}
Fix a distribution $\rho$ on $\NNI$.  The real users are assigned to batches by either of the basic schemes of Section~\ref{sec:defs}: under Poisson subsampling, user $i$ is included in batch $j$ according to an independent indicator $Z_{i,j}\sim\mathrm{Bern}(\gamma)$; under random allocation, user $i$ draws a uniformly random $k$-subset $A_i\subseteq[t]$ and is included in batch $j$ exactly when $j\in A_i$.  Whenever user $i$ is assigned to batch $j$, it sends its real contribution to that batch together with $D_{i,j}$ dummy elements $\bot$, where the dummy counts $D_{i,j}\sim\rho$ are drawn independently.

The server processes the batches sequentially.  If $U_j$ is the sequence of real users assigned to batch $j$ and $B_j$ is the resulting sequence after adding all participant-generated dummies, the released step-$j$ output is
\[
    y_j=(c_j,o_j),\qquad
    c_j:=|B_j|,\qquad
    o_j:=\alg(B_j,(y_1,\ldots,y_{j-1},c_j)).
\]
The transcript is $Y=(y_1,\ldots,y_t)$.  We denote the resulting mechanisms by $D_\rho^{\mathrm{Pois}}$ and $D_\rho^{\mathrm{RA}}$ for Poisson subsampling and random allocation, respectively.  Let $C_m:=\rho^{*m}$ be the distribution of $\sum_{\ell=1}^m D_\ell$, where the $D_\ell$ are i.i.d. from $\rho$ and $C_0$ is the point mass at zero.  We write $1+C_m$ for the distribution of $1+W$ when $W\sim C_m$.  If a batch contains $m$ common real participants, then its count with and without the distinguished user is distributed as $m+1+C_{m+1}$ and $m+C_m$ respectively, that is, as a common shift of the pair
\[
    (1+C_{m+1},\ C_m).
\]

\paragraph{The equivalent batch algorithm $M_\rho$.}
Define the batch algorithm $M_\rho=M_\rho[\alg]$ as follows.  On input a batch $U=(u_1,\ldots,u_m)$ and a previous view $v$, independently draw $D_1,\ldots,D_m\sim\rho$, form the batch
\[
    B=(u_1,\ldots,u_m,
      \underbrace{\bot,\ldots,\bot}_{D_1+\cdots+D_m\text{ copies}}),
\]
release the count
\[
    c:=m+\sum_{\ell=1}^m D_\ell,
\]
and then release
\[
    o:=\alg(B,(v,c)).
\]
Thus $M_\rho(U,v)=(c,o)$.  Unlike $M_C$ in Section~\ref{sec:batch}, the dummy distribution used by $M_\rho$ depends on the input size: if $|U|=m$, the dummy count has distribution $C_m$.

For a public target size $s$, the $s$-truncated variant $M_{\rho,s}$ generates the participant dummies and releases $c$ in the same way, then applies the truncation/padding map $T_{c,s}$ of Claim~\ref{clm:truncadj} before running $\alg$:
\[
  o:=\alg\bigl(T_{c,s}(B),(v,c)\bigr).
\]
 We write $D_{\rho,s}^{\mathrm{Pois}}$ and $D_{\rho,s}^{\mathrm{RA}}$ for the corresponding full transcripts.

\begin{lem}[Reduction to the participant-generated batch algorithm]\label{lem:participant-reduction}
For every dataset, as distributions over transcripts, the participant-generated versions of both sampling schemes are exactly the corresponding basic sampling schemes run on $M_\rho$:
\[
    D_\rho^{\mathrm{Pois}}(S)
      \ \stackrel{d}{=}\ \Pois{t,\gamma}{M_\rho}(S),
    \qquad
    D_\rho^{\mathrm{RA}}(S)
      \ \stackrel{d}{=}\ \alloc{t,k}{M_\rho}(S).
\]
\end{lem}
\begin{proof}
In either description the basic sampling scheme first determines the real users in every batch.  Conditional on these assignments, applying $M_\rho$ to a batch draws, independently for every real participant in that batch, a $\rho$-distributed number of dummy elements, releases the resulting count, and then applies $\alg$ with the same adaptive view and count.  This is exactly the participant-side dummy generation and server-side release rule above, step by step.
\end{proof}

The same reduction holds for the fixed-size versions after replacing $M_\rho,D_\rho^{\mathrm{Pois}},D_\rho^{\mathrm{RA}}$ by $M_{\rho,s},D_{\rho,s}^{\mathrm{Pois}},D_{\rho,s}^{\mathrm{RA}}$, respectively.

The following monotonicity property replaces the fixed count pair $(C_{+1},C)$ of Section~\ref{sec:batch}: additional common participants can only help, since they add the same independent noise to both neighboring count distributions.

\begin{lem}[Monotonicity of the participant count pair]\label{lem:participant-count-envelope}
For every pair of integers $m\ge r\ge0$, the pair
\[
    (1+C_{r+1},C_r)
\]
dominates $(1+C_{m+1},C_m)$.  Equivalently, the swapped pair $(C_r,1+C_{r+1})$ dominates $(C_m,1+C_{m+1})$.
\end{lem}
\begin{proof}
Let $X\sim C_{r+1}$, $Y\sim C_r$, and let $W\sim C_{m-r}$ be drawn independently. Then $1+X+W\sim1+C_{m+1}$ and $Y+W\sim C_m$.
Thus $(1+C_{m+1},C_m)$ is obtained from $(1+C_{r+1},C_r)$ by the same Markov kernel, namely adding an independent $C_{m-r}$ draw.  Hockey-stick divergence contracts under post-processing, proving the remove-direction statement.  The swapped statement follows by the same argument with the two distributions interchanged.
\end{proof}

\begin{lem}[Dominating pair for a large participant-generated batch]\label{lem:participant-batchpair}
Fix $r\ge0$ and suppose the core mechanism $\alg$ is dominated by $(P_0,Q_0)$ in the remove direction of the zero-out adjacency. Define
\[
    P_{\rho,r}:=(1+C_{r+1})\times P_0,
    \qquad
    Q_{\rho,r}:=C_r\times Q_0 .
\]
Then for every view $v$ and every pair of add/remove neighboring batches $U,U_-$, where $U_-$ is obtained from $U$ by deleting one element $x_n$ and $|U_-|\ge r$, the pair $(M_\rho(U,v),M_\rho(U_-,v))$ is dominated by $(P_{\rho,r},Q_{\rho,r})$.  Consequently $(M_\rho(U_-,v),M_\rho(U,v))$ is dominated by the swapped pair $(Q_{\rho,r},P_{\rho,r})$.
\end{lem}
\begin{proof}
Fix $v$ and write $M_\rho(\cdot)=M_\rho(\cdot,v)$.  As in Lemma~\ref{lem:batchpair-kb}, expose the output in two stages: first the count $c$, then the core output $o$ given $c$.  Let $a:=|U_-|$, so $a\ge r$ and $U$ has $a+1$ real elements.  Write $K_1(W)$ for the distribution of the released count and $K_2(W,c)$ for the conditional distribution of the core output given count $c$.

\emph{Count.}  Under $U_-$ the dummy count is distributed as $C_a$, while under $U$ it is distributed as $C_{a+1}$.  Hence $K_1(U)$ is the distribution of $a+1+W$ for $W\sim C_{a+1}$, and $K_1(U_-)$ the distribution of $a+W$ for $W\sim C_a$.
Translation by $-a$ is a bijection and preserves every hockey-stick divergence, so the count stage is dominated by $(1+C_{a+1},C_a)$.  Since $a\ge r$, Lemma~\ref{lem:participant-count-envelope} further dominates this pair by $(1+C_{r+1},C_r)$.

\emph{Output given count.}  Fix a count $c$.  If $c<a+1$, then this count is unattainable under $U$; redefining $K_2(U,c):=K_2(U_-,c)$ there makes the conditional pair identical and does not change the distribution of $M_\rho(U)$.  Now assume $c\ge a+1$.  Conditional on the released count $c$, the batch passed to $\alg$ under $U$ consists of the $a$ common real elements, the distinguished element $x_n$, and $c-a-1$ dummies.  Under $U_-$, use the corresponding size-$c$ batch consisting of the same $a$ common real elements and $c-a$ dummies; when the count $c$ is attainable under $U_-$ this is the true conditional batch, and otherwise it is a harmless definition of the off-support kernel.  These two size-$c$ batches differ in one entry, $x_n$ versus $\bot$, and $\alg$ receives the same view-with-count $(v,c)$ in both worlds.  By the hypothesis on $\alg$, the conditional pair $(K_2(U,c),K_2(U_-,c))$ is dominated by $(P_0,Q_0)$ for every $c$.

\emph{Compose.}  By the adaptive composition theorem for dominating pairs,
\[
    \bigl(M_\rho(U),M_\rho(U_-)\bigr)
    \quad\text{is dominated by}\quad
    (1+C_{r+1},C_r)\times(P_0,Q_0)
    =(P_{\rho,r},Q_{\rho,r}).
\]
Since the statement holds for every neighboring pair satisfying $|U_-|\ge r$ and every view, the swap fact of Section~\ref{sec:defs} gives the add direction.
\end{proof}

The conclusion of Lemma~\ref{lem:participant-batchpair} also holds with $M_\rho$ replaced by $M_{\rho,s}$, by Claim~\ref{clm:truncadj} and Lemma~\ref{lem:adjpres} exactly as in the proof of Lemma~\ref{lem:batchpair}. Accordingly, every amplification statement below has an $s$-truncated counterpart obtained by replacing $D_\rho$ with $D_{\rho,s}$.

Lemma~\ref{lem:participant-batchpair} is the batch-level statement needed for either amplification result.  The remaining issue is that neither sampling scheme guarantees $|U_-^j|\ge r$ in every batch; the failure probability of this minimum-participation event is charged to $\delta$.

\subsection{Amplification under a minimum-participation event}\label{sec:participant-amplification}

For add/remove neighbors $S=(x_1,\dots,x_n)$ and $S_-=(x_1,\dots,x_{n-1})$ differing in the distinguished user $x_n$, let $N_j$ denote the number of common users (users of $S_-$) assigned to batch $j$, before dummies are added, and let $E_r$ be the good event that every batch contains at least $r$ common users. Under both sampling schemes each $N_j$ is distributed as $\mathrm{Binomial}(n-1,p)$, with $p=\gamma$ for Poisson subsampling and $p=k/t$ for random allocation; the $N_j$ are independent under Poisson subsampling and dependent under random allocation. Writing $\mu=(n-1)p$, if $r\le(1-\eta)\mu$ for some $\eta\in[0,1]$, the multiplicative Chernoff bound together with a union bound over the $t$ batches gives
\begin{equation}\label{eq:participant-beta-ra-chernoff}
    \Pr[\,E_r\,]
      \ge 1-t\exp\!\left(-\frac{\eta^2\mu}{2}\right).
\end{equation}

\begin{cor}[Random allocation with participant-generated count noise]\label{cor:participant-ra}
Let $D_\rho^{\mathrm{RA}}(S)=\alloc{t,k}{M_\rho}(S)$ denote the full transcript, including both the counts and the outputs of $\alg$.  Fix $r\ge0$ and suppose $\Pr[E_r]\ge 1-\beta$.  Let%
\[
    \mathsf P_{\rho,r}^{\mathrm{RA}}
      :=\mathsf{RA}_{t,k}(P_{\rho,r},Q_{\rho,r}),
    \qquad
    \mathsf Q_{\rho,r}^{\mathrm{RA}}
      :=Q_{\rho,r}^{\otimes t},
\]
where $P_{\rho,r},Q_{\rho,r}$ are as in Lemma~\ref{lem:participant-batchpair}.  Then for every $\kappa\ge0$,
\begin{align}
    \HockeyStickDiv{\kappa}{D_\rho^{\mathrm{RA}}(S)}{D_\rho^{\mathrm{RA}}(S_-)}
    &\le \beta+(1-\beta)\,
      \HockeyStickDiv{\kappa}{\mathsf P_{\rho,r}^{\mathrm{RA}}}
                              {\mathsf Q_{\rho,r}^{\mathrm{RA}}},
    \label{eq:participant-ra-remove}\\
    \HockeyStickDiv{\kappa}{D_\rho^{\mathrm{RA}}(S_-)}{D_\rho^{\mathrm{RA}}(S)}
    &\le \beta+(1-\beta)\,
      \HockeyStickDiv{\kappa}{\mathsf Q_{\rho,r}^{\mathrm{RA}}}
                              {\mathsf P_{\rho,r}^{\mathrm{RA}}}.
    \label{eq:participant-ra-add}
\end{align}
Consequently, with
\[
    \delta_{\rho,r}^{\mathrm{RA}}(\varepsilon)
      :=\max\left\{
        \HockeyStickDiv{e^\varepsilon}{\mathsf P_{\rho,r}^{\mathrm{RA}}}
                                      {\mathsf Q_{\rho,r}^{\mathrm{RA}}},
        \HockeyStickDiv{e^\varepsilon}{\mathsf Q_{\rho,r}^{\mathrm{RA}}}
                                      {\mathsf P_{\rho,r}^{\mathrm{RA}}}
      \right\},
\]
the participant-generated random-allocation release is
\[
    \left(\varepsilon,\;\beta+
      (1-\beta)\,\delta_{\rho,r}^{\mathrm{RA}}(\varepsilon)\right)\text{-DP}.
\]
\end{cor}
\begin{proof}
First condition on a fixed allocation of the $n-1$ common users for which $E_r$ holds; then every batch of $S_-$ has at least $r$ real participants. Given this allocation, the transcript is the $k$-out-of-$t$ random allocation of the distinguished user alone. In each of the $k$ batches that the distinguished user joins, the two neighboring inputs to $M_\rho$ are $U^j$ and $U_-^j$ with $|U_-^j|\ge r$, so by Lemma~\ref{lem:participant-batchpair} the conditional output pair is dominated by $(P_{\rho,r},Q_{\rho,r})$ for every adaptive view; in the remaining batches the two inputs coincide, and the pair is trivially dominated by $(Q_{\rho,r},Q_{\rho,r})$. The proof of Lemma~\ref{lem:RAdom} in \citep{FS25ra} conditions on the allocation of the remaining users in exactly this way and uses only the per-step domination of the batches that actually arise; it therefore applies verbatim and gives domination by $(\mathsf P_{\rho,r}^{\mathrm{RA}},\mathsf Q_{\rho,r}^{\mathrm{RA}})$ conditional on this common allocation.

Averaging over the common-user allocation conditional on $E_r$ is a mixture with the \emph{same} weights in both worlds, because that conditional allocation has the same distribution under $S$ and $S_-$ and is independent of the distinguished user's allocation; joint convexity of the hockey-stick divergence therefore shows that the transcript pair conditional on $E_r$ is dominated by $(\mathsf P_{\rho,r}^{\mathrm{RA}},\mathsf Q_{\rho,r}^{\mathrm{RA}})$. Finally, $E_r$ depends only on the common users, so it has the same probability $b\ge1-\beta$ under $S$ and $S_-$. Decomposing both transcript distributions according to $E_r$, joint convexity together with the elementary bound $\HockeyStickDiv{\kappa}{P}{Q}\le1$ on the complement of $E_r$ gives

\[
    \HockeyStickDiv{\kappa}{D_\rho^{\mathrm{RA}}(S)}{D_\rho^{\mathrm{RA}}(S_-)}
    \le (1-\beta)\,
      \HockeyStickDiv{\kappa}{\mathsf P_{\rho,r}^{\mathrm{RA}}}
                              {\mathsf Q_{\rho,r}^{\mathrm{RA}}}
      +\beta.
\]
The reverse direction is identical with the pair swapped.  Setting $\kappa=e^\varepsilon$ proves the DP statement.
\end{proof}

\begin{cor}[Poisson subsampling with participant-generated count noise]\label{cor:participant-poisson}
Let $D_\rho^{\mathrm{Pois}}(S)=\Pois{t,\gamma}{M_\rho}(S)$ denote the full $t$-step transcript, including both the counts and the outputs of $\alg$.  Fix $r\ge0$ and suppose $\Pr[E_r]\ge 1-\beta$.  Define the one-step amplified pair
\[
    P_{\rho,r,\gamma}
      :=\gamma P_{\rho,r}+(1-\gamma)Q_{\rho,r},
\]
and its $t$-fold composition
\[
    \mathsf P_{\rho,r}^{\mathrm{Pois}}
      :=P_{\rho,r,\gamma}^{\otimes t},
    \qquad
    \mathsf Q_{\rho,r}^{\mathrm{Pois}}
      :=Q_{\rho,r}^{\otimes t}.
\]
Then for every $\kappa\ge0$,
\begin{align}
    \HockeyStickDiv{\kappa}{D_\rho^{\mathrm{Pois}}(S)}{D_\rho^{\mathrm{Pois}}(S_-)}
    &\le \beta+(1-\beta)\,
      \HockeyStickDiv{\kappa}{\mathsf P_{\rho,r}^{\mathrm{Pois}}}
                              {\mathsf Q_{\rho,r}^{\mathrm{Pois}}},
    \label{eq:participant-poisson-remove}\\
    \HockeyStickDiv{\kappa}{D_\rho^{\mathrm{Pois}}(S_-)}{D_\rho^{\mathrm{Pois}}(S)}
    &\le \beta+(1-\beta)\,
      \HockeyStickDiv{\kappa}{\mathsf Q_{\rho,r}^{\mathrm{Pois}}}
                              {\mathsf P_{\rho,r}^{\mathrm{Pois}}}.
    \label{eq:participant-poisson-add}
\end{align}
Consequently, with
\[
    \delta_{\rho,r}^{\mathrm{Pois}}(\varepsilon)
      :=\max\left\{
        \HockeyStickDiv{e^\varepsilon}{\mathsf P_{\rho,r}^{\mathrm{Pois}}}
                                      {\mathsf Q_{\rho,r}^{\mathrm{Pois}}},
        \HockeyStickDiv{e^\varepsilon}{\mathsf Q_{\rho,r}^{\mathrm{Pois}}}
                                      {\mathsf P_{\rho,r}^{\mathrm{Pois}}}
      \right\},
\]
the participant-generated Poisson-subsampling release is
\[
    \left(\varepsilon,\;\beta+
      (1-\beta)\,\delta_{\rho,r}^{\mathrm{Pois}}(\varepsilon)\right)\text{-DP}.
\]
For one step, the good-event pair is simply $(P_{\rho,r,\gamma},Q_{\rho,r})$, exactly the Poisson-amplified pair of Lemma~\ref{lem:subsamPLDdom}.
\end{cor}
\begin{proof}
Condition on the inclusion indicators of the common users, for a realization in $E_r$.  At every step $j$ the batch of $S_-$ then has $|U_-^j|\ge r$ real participants, and the distinguished user joins it independently with probability $\gamma$.  By Lemma~\ref{lem:participant-batchpair}, the pair $(M_\rho(U_-^j\cup\{x_n\}),M_\rho(U_-^j))$ is dominated by $(P_{\rho,r},Q_{\rho,r})$ for every view, so the argument of Lemma~\ref{lem:subsamPLDdom} gives the one-step pair $(P_{\rho,r,\gamma},Q_{\rho,r})$; adaptive composition over the $t$ steps gives the stated product pair.  Averaging over the common-user samples in $E_r$ and then charging its complement proceeds exactly as in Corollary~\ref{cor:participant-ra}.
\end{proof}

\paragraph{Examples for the count coordinate.}
If $\rho=\mathrm{Bern}(q)$, then the count coordinates of the batch pair are
\[
    Q_{\rho,r}^{\mathrm{cnt}}=\mathrm{Binomial}(r,q),
    \qquad
    P_{\rho,r}^{\mathrm{cnt}}=1+\mathrm{Binomial}(r+1,q).
\]
Since $\mathrm{Binomial}(r+1,q)$ is the mixture, with weights $1-q$ and $q$, of $\mathrm{Binomial}(r,q)$ and $1+\mathrm{Binomial}(r,q)$, the count pair is in this case a mixture of the sensitivity-one pair $(1+\mathrm{Binomial}(r,q),\mathrm{Binomial}(r,q))$ and the sensitivity-two pair $(2+\mathrm{Binomial}(r,q),\mathrm{Binomial}(r,q))$ of binomial count noise.

\section{The cost of masking: count noise under subsampling and composition}\label{sec:onesided}

The computational and communication overhead of masking is determined by the expected number of dummies per batch. We first analyze this overhead in isolation, as if only the masked counts were released, and then, in Section~\ref{sec:onesided-applications}, examine the number of dummies needed when masking the counts of a Gaussian per-batch mechanism.

The privacy guarantees in Sections~\ref{sec:batch} and~\ref{sec:amp} reduce the contribution of the released counts to a simple pair of distributions. If the dummy count has distribution $C$ on $\NNI$, then changing one user's participation shifts that distribution by one, giving the pair $(C_{+1},C)$. Under Poisson subsampling at rate $\gamma$, this pair becomes
\[
    P_\gamma := (1-\gamma)C+\gamma C_{+1},
    \qquad
    Q:=C,
\]
and the $t$ releases are accounted for through $(P_\gamma^{\otimes t},Q^{\otimes t})$ and the swapped pair. In isolation, then, the question is how small the mean $\E_{R\sim C}[R]$ can be while both composed pairs satisfy the target guarantee. Writing
\[
    \delta_C^{\gamma,t}(\eps)
    :=
    \max\left\{
        \HockeyStickDiv{e^\eps}{P_\gamma^{\otimes t}}{Q^{\otimes t}},
        \HockeyStickDiv{e^\eps}{Q^{\otimes t}}{P_\gamma^{\otimes t}}
    \right\}
\]
for the privacy profile of the count release in both orientations, the quantity of interest is
\begin{equation}\label{def:onesided-problem}
    \meanopt(\eps,\delta;\gamma,t)
    :=
    \inf\left\{
        \E_{R\sim C}[R]
        : C\text{ is supported on }\NNI,
          \ \delta_C^{\gamma,t}(\eps)\le \delta
    \right\}.
\end{equation}

The focus of this section is on asymptotics, and we will assume throughout that
\begin{align*}
    \gamma t\ge1,
    \qquad
    \logdelta\ge1,
    \qquad
    \eps\le \logdelta,
    \qquad
    \sigmastar\to\infty,
    \tag{A}\label{eq:onesided-asymptotics}
\end{align*}
where
\[
    \logdelta:=\log\frac1\delta,
    \qquad
    \sigmastar:=\frac{\gamma\sqrt{2t\logdelta}}\eps .
\]
The first three conditions are standard normalizations that hold in any regime of interest; the last is the standard assumption in asymptotic analyses of subsampled composition \citep{ACGMMT16,FS25ra}. Note that $\sigmastar$ is a natural point of comparison, since up to constants it is the noise multiplier of a $\gamma$-subsampled, $t$-fold composed Gaussian mechanism at the same $(\eps,\delta)$ \citep{ACGMMT16,Steinke22}.

We prove two upper bounds on the smallest expected dummy count (eq.~\eqref{def:onesided-problem}):
\[
    \meanopt(\eps,\delta;\gamma,t)
    = O\!\left(\sigmastar\log\frac{\gamma t}{\delta}\right)
    \quad\text{(Proposition~\ref{prop:onesided-blackbox})},
    \qquad
    \meanopt(\eps,\delta;\gamma,t)
    = O\!\left(\sigmastar\log\sigmastar\right)
    \quad\text{(Theorem~\ref{thm:onesided-clt})}.
\]
The first applies standard subsampling and composition theorems to a private one-shot count release. The second holds under a CLT condition (Definition~\ref{def:onesided-clt} below), which asserts that the composition has the $(\eps,\delta)$ guarantee of Gaussian differential privacy with the parameter determined by the $\chi^2$ divergence of a single step. Both bounds rely on a truncated two-sided geometric noise distribution with the same width parameter, and differ only in the mean. The black-box route requires boundary mass at most $\delta/(\gamma t)$, placing the center at value $\Theta(\sigmastar\log\frac{\gamma t}\delta)$; under the CLT condition the boundary mass only has to be comparable to the divergence budget $1/{\sigmastar}^2$, for which distance $O(\sigmastar\log\sigmastar)$ suffices. The second bound is thus the smaller one whenever $\log\sigmastar\ll\log\frac{\gamma t}\delta$, which is the case in most practical regimes where  amplification is used and $t$ is large. The proofs omitted from this section appear in Appendix~\ref{app:onesided}.

\subsection{The count noise family}\label{sec:onesided-standalone}

We start with some intuition on non-negative noise distributions. When $\gamma=t=1$ the neighboring distributions are $C_{+1}$ and $C$. In the add direction, $C$ assigns probability $C(0)$ to zero while $C_{+1}$ assigns none, so any $(\eps,\delta)$ guarantee requires $C(0)\le\delta$; the $\eps$ constraint further limits how fast the probabilities can grow away from zero. Known results on one-sided count noise make this exact: \citet{CHM21} give a truncated geometric construction for the problem, while exact global optimality is established in \cite[Theorem~1]{XWSD25}. For $\eps\le1$ and $\delta<\eps/2$ the resulting optimum is
\begin{equation}\label{eq:onesided-standalone-cost}
    \meanopt(\eps,\delta;1,1)
    =\Theta\!\left(\frac1\eps\log\frac{\eps}\delta\right),
\end{equation}
attained by a shifted, clamped two-sided geometric distribution.
Both our bounds are attained by such a family: a two-sided geometric distribution of width $b$, recentered at $m$ and truncated at $0$. They differ only in the calibration of $m$.

\begin{lem}[Recentered geometric count noise]\label{lem:onesided-geometric}
For $b\ge2$ and an integer $m\ge0$, let
\[
    z:=1-\frac1b,
    \qquad
    C_{b,m}(k):=\frac{z^{|k-m|}}{Z},
    \qquad
    Z:=\sum_{j\ge0}z^{|j-m|},
\]
and write $C=C_{b,m}$. Then:
\begin{enumerate}
\item\label{item:onesided-shape}\emph{(Shape and mean.)} $Z\in[b,2b]$, $\ C(0)\le e^{-m/b}/b$, and $m\le\E_{R\sim C}[R]\le m+2b$.
\item\label{item:onesided-ratio}\emph{(Likelihood ratio.)} $C_{+1}/C$ equals $0$ at $k=0$, $z$ for $1\le k\le m$, and $z^{-1}$ for $k>m$.
\item\label{item:onesided-dp}\emph{(One-shot privacy.)} $\delta_C^{1,1}(\eps_0)=C(0)$ for $\eps_0:=\log\frac1z$; more precisely, $\HockeyStickDiv{e^{\eps_0}}{C_{+1}}{C}=0$ and $\HockeyStickDiv{e^{\eps_0}}{C}{C_{+1}}=C(0)$.
\item\label{item:onesided-chi}\emph{($\chi^2$ divergence.)} $\chi^2(C_{+1}\|C)\le C(0)+4/b^2$.
\end{enumerate}
\end{lem}

We use the $\chi^2$ divergence because it controls the first two moments of the privacy loss random variable \cite[Lemma~4.1]{FS26clt}; recall that it is the likelihood-ratio second moment minus one:
\begin{equation}\label{eq:onesided-chi-moment}
      \chi^2(P\|Q)
      =\E_{\omega\sim Q}\!\left[\left(\frac{P(\omega)}{Q(\omega)}-1\right)^{2}\right]
      =\E_{\omega\sim Q}\!\left[\left(\frac{P(\omega)}{Q(\omega)}\right)^{2}\right]-1 .
\end{equation}

The proof of Lemma~\ref{lem:onesided-geometric} is an elementary computation given in Appendix~\ref{app:onesided-geometric}. We use two calibrations of the center $m$: for a parameter $\delta_0\in(0,1)$,
\begin{equation}\label{eq:onesided-calibrations}
    m=\left\lceil b\log\frac1{b\delta_0}\right\rceil_+
    \ \Longrightarrow\
    C(0)\le\delta_0,
    \qquad\qquad
    m=\left\lceil b\log b\right\rceil
    \ \Longrightarrow\
    \chi^2(C_{+1}\|C)\le\frac5{b^2},
\end{equation}
where $\lceil x\rceil_+:=\max\{0,\lceil x\rceil\}$; both are immediate from parts \eqref{item:onesided-shape} and \eqref{item:onesided-chi}, the second using $C(0)\le1/b^2$. By part~\eqref{item:onesided-shape} the resulting means are at most $b\bigl[\log\frac1{b\delta_0}\bigr]_++2b+1$ and $b\log b+2b+1$, where $[x]_+:=\max\{0,x\}$.

\subsection{Count noise under subsampling and composition}
We now describe the two upper bounds on the smallest expected dummy count (eq.~\eqref{def:onesided-problem}).
\paragraph{A black-box upper bound.}
We use the first calibration of \eqref{eq:onesided-calibrations}, with $\eps_0=1/(8\sigmastar)$ and $\delta_0=\delta/(2\gamma t)$, and then apply standard bounds on privacy amplification by Poisson subsampling \citep{BBG18,Steinke22} and advanced composition \citep{DRV10,DR14} to the resulting pair. The details are in Appendix~\ref{app:onesided-blackbox}.

\begin{prop}[Black-box upper bound]\label{prop:onesided-blackbox}
Assume the asymptotic conditions~\eqref{eq:onesided-asymptotics}. Then $\meanopt(\eps,\delta;\gamma,t)=O\bigl(\sigmastar\log\frac{\gamma t}{\delta}\bigr)$.
\end{prop}

We now argue that the extra logarithmic factor is an artifact of black-box composition and can be removed by a more careful analysis.
\paragraph{The CLT regime.}
The privacy loss random variable of the pair $(P_\gamma^{\otimes t},Q^{\otimes t})$ is the sum of $t$ independent copies of the privacy loss of the single-step pair $(P_\gamma,Q)$. The central limit theorem (CLT) for privacy loss distributions \citep{SMM18,BDLS20,DRS22} predicts that such a composition is approximately Gaussian differentially private (GDP). For $V\ge0$ we denote by $\eps_{\mathrm G}(V,\delta)$ the smallest $\eps\ge0$ such that $\sqrt V$-GDP implies $(\eps,\delta)$-DP, that is, such that $\HockeyStickDiv{e^\eps}{N(\sqrt V,1)}{N(0,1)}\le\delta$. By \cite[Lemma~12 and Remark~15]{Steinke22} (with $\rho=V/2$), we have that
\begin{equation}\label{eq:onesided-gdp-tail}
    \eps_{\mathrm G}(V,\delta)\ \le\ \frac{V}{2}+\sqrt{2V\logdelta}.
\end{equation}
This bound is tight up to lower-order terms when $V$ is not too small compared with $\delta$. Specifically, $\eps_{\mathrm G}(V,\delta)=V/2+(1+o(1))\sqrt{2V\logdelta}$ as $\delta\to0$ and $V\ge\delta^{o(1)}$ (Lemma~\ref{lem:onesided-gdp-asymp}).
\citet{FS26clt} show that this Gaussian approximation holds up to relatively small corrections and that its parameter is determined by the $\chi^2$ divergence $v$ of the pair that dominates a single step. Specifically, the privacy loss of $t$ steps of Poisson subsampling at rate $\gamma$ is close to that of $\sqrt{t\gamma^2v/(1-\gamma)}$-GDP \cite[Corollaries~4.5 and~4.7]{FS26clt}, and that of $1$-out-of-$t$ random allocation is close to that of $\sqrt{v/t}$-GDP \cite[Theorem~5.2]{FS26clt}. These results hold when the privacy loss of each step is small except with small probability. The corrections are a shift of $\eps$ that is of lower order in typical applications, an additive term in $\delta$ and, for Poisson subsampling, a relative error in the variance of the privacy loss that is of order $\gamma$ for a fixed pair. To keep this asymptotic presentation simple, we ignore these corrections and assume that the approximation is exact, in the following form.

\begin{defn}[CLT condition]\label{def:onesided-clt}
Let $(P_0,Q_0)$ be a pair of distributions with $v:=\chi^2(P_0\|Q_0)<\infty$, let $t\ge1$ be an integer, and let $\delta\in(0,1)$. For $\gamma\in(0,1)$ and $P_{0,\gamma}:=(1-\gamma)Q_0+\gamma P_0$, we say that $(P_0,Q_0)$ satisfies the \emph{CLT condition} for Poisson subsampling at rate $\gamma$, $t$ and $\delta$ if the smallest $\eps\ge0$ for which
\[
    \max\left\{\HockeyStickDiv{e^\eps}{P_{0,\gamma}^{\otimes t}}{Q_0^{\otimes t}},\ \HockeyStickDiv{e^\eps}{Q_0^{\otimes t}}{P_{0,\gamma}^{\otimes t}}\right\}\ \le\ \delta
\]
is equal to $\eps_{\mathrm G}\bigl(t\gamma^2v/(1-\gamma),\delta\bigr)$. Similarly, $(P_0,Q_0)$ satisfies the \emph{CLT condition} for random allocation, $t$ and $\delta$ if the same holds for the pair $(\mathsf{RA}_{t,1}(P_0,Q_0),Q_0^{\otimes t})$ of Definition~\ref{def:RApair} in place of $(P_{0,\gamma}^{\otimes t},Q_0^{\otimes t})$, with $\eps_{\mathrm G}(v/t,\delta)$ in place of $\eps_{\mathrm G}(t\gamma^2v/(1-\gamma),\delta)$.
\end{defn}
Note that in both cases the variance of the Gaussian approximation is proportional to $v$. For our upper bound in Theorem~\ref{thm:onesided-clt} the simplifying exactness of the CLT condition is used primarily for brevity as the corrections listed above do not affect the asymptotic conclusion. In Section~\ref{sec:onesided-applications} we use the CLT condition to compare privacy bounds using the corresponding $\chi^2$ divergences. There we effectively assume that the differences in the $\chi^2$ divergences are not outweighed by the differences in correction terms.

Under the CLT condition the width of count noise needs to be at the Gaussian scale $b=\Theta(\sigmastar)$ while the second calibration then requires the center at $b\log b$ (which is where the remaining logarithmic factor in the mean comes from).
\begin{thm}[Count noise in the CLT regime]\label{thm:onesided-clt}
Assume the asymptotic conditions~\eqref{eq:onesided-asymptotics} and $\gamma\le\frac12$. Let
\[
    b:=\lceil 4\sigmastar\rceil,
    \qquad
    m:=\lceil b\log b\rceil,
    \qquad
    C:=C_{b,m}.
\]
If $(C_{+1},C)$ satisfies the CLT condition for Poisson subsampling at rate $\gamma$, $t$ and $\delta$, then $\delta_C^{\gamma,t}(\eps)\le\delta$ and
\[
    \E_{R\sim C}[R]\ \le\ b\log b+2b+1 .
\]
In particular,
\[
    \meanopt(\eps,\delta;\gamma,t)=O\!\left(\sigmastar\log\sigmastar\right).
\]
\end{thm}
\begin{proof}
Note that $b\ge2$ for all sufficiently large parameters by the asymptotic conditions~\eqref{eq:onesided-asymptotics}, so the second calibration of the center $m$ (eq.~\eqref{eq:onesided-calibrations}) gives $v:=\chi^2(C_{+1}\|C)\le5/b^2$. Let $V:=t\gamma^2v/(1-\gamma)$ be the variance in the CLT condition. Since $\gamma\le\frac12$, $b\ge4\sigmastar$ and ${\sigmastar}^2=2\logdelta t\gamma^2/\eps^2$, we have that
\[
    V\ \le\ \frac{10\,t\gamma^2}{b^2}
       \ \le\ \frac{10\,t\gamma^2}{16{\sigmastar}^2}
       \ =\ \frac{5\,\eps^2}{16\logdelta}.
\]
Therefore, by the tail bound on $\eps_{\mathrm G}$ (eq.~\eqref{eq:onesided-gdp-tail}) and $\eps\le\logdelta$,
\[
    \eps_{\mathrm G}(V,\delta)
      \ \le\ \frac{V}{2}+\sqrt{2V\logdelta}
      \ \le\ \frac{5\,\eps}{32}+\sqrt{\frac58}\,\eps
      \ \le\ \eps .
\]
Applying the CLT condition to $(C_{+1},C)$ therefore bounds both hockey-stick divergences in the definition of $\delta_C^{\gamma,t}(\eps)$ by $\delta$. The mean bound is Lemma~\ref{lem:onesided-geometric}\eqref{item:onesided-shape} together with $m\le b\log b+1$, and $b=\Theta(\sigmastar)$ turns it into $O(\sigmastar\log\sigmastar)$.
\end{proof}

\subsection{Masking the count of a Gaussian mechanism}\label{sec:onesided-applications}

Theorem~\ref{thm:onesided-clt} bounds the count noise in terms of $\sigmastar$, as if only the masked counts were released. In an application, however, the count accompanies the output of a per-batch mechanism that already consumes the privacy budget. We therefore express the count noise mean in terms of the parameters of the typical application: the noise multiplier $\sigma$ of the per-batch Gaussian mechanism and the fraction $\alpha$ by which one is willing to let $\eps$ grow. Specifically, we show that, for both Poisson subsampling and random allocation, count noise of mean $O(\frac{\sigma}{\sqrt\alpha}\log\frac{\sigma}{\sqrt\alpha})$ suffices to ensure that the privacy guarantee of the Gaussian mechanism degrades by at most a factor of $1+\alpha$.

More formally, we consider a per-batch Gaussian mechanism with noise multiplier $\sigma$: the standard deviation of the added noise is $\sigma$ times the $\ell_2$ sensitivity of the sum being computed. This covers both of our applications: the noisy sums of clipped gradients in DP-SGD, and the noisy aggregates of distributed mean estimation (Section~\ref{sec:distributed}), where the observable count is the number of messages in an aggregate. Its one-dimensional dominating pair is $P_0=N(1,\sigma^2)$, $Q_0=N(0,\sigma^2)$, whose likelihood ratio has $Q_0$-second moment exactly $e^{1/\sigma^2}$, so
\begin{equation}\label{eq:onesided-gauss-chi}
    \chi_s^2:=\chi^2(P_0\|Q_0)=e^{1/\sigma^2}-1=\Theta(\sigma^{-2}).
\end{equation}
We assume that the scheme is calibrated so that the Gaussian mechanism alone, when subsampled at rate $\gamma$ and composed $t$ times (or under $1$-out-of-$t$ random allocation), satisfies $(\eps,\delta)$ for some $\eps<1$, and we ask that masking the count degrade this to no worse than $\bigl((1+\alpha)\eps,\delta\bigr)$.

In the notation of Section~\ref{sec:batch}, each batch algorithm (before subsampling) is dominated by the pair $P=C_{+1}\times P_0$, $Q=C\times Q_0$, a product pair, so by the definition of the $\chi^2$ divergence (eq.~\eqref{eq:onesided-chi-moment}) its likelihood-ratio second moment factorizes: with $\chi_c^2:=\chi^2(C_{+1}\|C)$,
\[
    1+\chi^2(P\|Q)=(1+\chi_c^2)(1+\chi_s^2).
\]
Therefore $\chi^2(P\|Q)=\chi_s^2+\chi_c^2(1+\chi_s^2)$, that is, masking increases the $\chi^2$ divergence of the pair before subsampling by the factor $1+\chi_c^2(1+\chi_s^2)/\chi_s^2$. Under the CLT condition the variance of the Gaussian approximation is proportional to the $\chi^2$ divergence of the pair for both Poisson subsampling and random allocation. Thus masking increases this variance by the same factor. When the bound on $\eps_{\mathrm G}$ (eq.~\eqref{eq:onesided-gdp-tail}) is tight up to lower-order terms, $\eps$ then grows by at most this factor, up to lower-order terms. Calibrating the count noise as in Theorem~\ref{thm:onesided-clt}, with width of order $\sigma/\sqrt\alpha$, makes this factor at most $1+\alpha$ and the mean of the count noise $O(\frac{\sigma}{\sqrt\alpha}\log\frac{\sigma}{\sqrt\alpha})$. The following theorem, proved in Appendix~\ref{app:onesided-gaussian}, makes this precise.
\begin{thm}[Masking the count of a Gaussian mechanism]\label{thm:onesided-gaussian}
Let $\sigma\ge1$ and $\alpha\in(0,1]$, and let
\[
    b:=\left\lceil\sigma\sqrt{80/\alpha}\right\rceil,
    \qquad
    m:=\lceil b\log b\rceil,
    \qquad
    C:=C_{b,m}.
\]
Consider either Poisson subsampling at rate $\gamma\le\frac12$ over $t$ steps, with $V_0:=t\gamma^2\chi_s^2/(1-\gamma)$, or $1$-out-of-$t$ random allocation, with $V_0:=\chi_s^2/t$ and $\gamma:=1/t$. Suppose that both $(P_0,Q_0)$ and $(P,Q)$ satisfy the CLT condition for this scheme, $t$ and $\delta$, and let $\eps:=\eps_{\mathrm G}(V_0,\delta)$, so that the unmasked release is $(\eps,\delta)$-DP. Assume the asymptotic conditions~\eqref{eq:onesided-asymptotics}, and that $\delta\to0$ with $V_0\ge\delta^{o(1)}$. Then, for a fixed $\alpha$ and all sufficiently large parameters, the masked release is $\bigl((1+\alpha)\eps,\delta\bigr)$-DP in both directions, at a cost of
\[
    \E_{R\sim C}[R]\ \le\ b\log b+2b+1
    \ =\ O\!\left(\frac{\sigma}{\sqrt\alpha}\,\log\frac{\sigma}{\sqrt\alpha}\right)
\]
dummies per batch. Equivalently, since $\sigma=\Theta(\sigmastar)$,
\[
    \E_{R\sim C}[R]
    \ =\ O\!\left(\frac{\sigmastar}{\sqrt\alpha}\,\log\frac{\sigmastar}{\sqrt\alpha}\right),
    \qquad
    \sigmastar=\frac{\gamma\sqrt{2t\log(1/\delta)}}{\eps}.
\]
\end{thm}
The last condition holds whenever $V_0$ is bounded below by a positive constant, and otherwise says that $\delta$ decreases faster than any fixed power of $V_0$. It restricts the theorem to the regime in which the bound on $\eps_{\mathrm G}$ (eq.~\eqref{eq:onesided-gdp-tail}) is tight up to lower-order terms (Lemma~\ref{lem:onesided-gdp-asymp}).
\section{Numerical Evaluation}\label{sec:experiments}

We evaluate dummy count masking numerically, using PLD accounting libraries for composition, subsampling and random allocation \citep{Google20,PLD_acc_msfft}. We focus on two questions. First, \emph{how much does dummy count masking increase the privacy parameter $\eps$}, and which of the count-noise families below gives the smallest overhead? We consider both independent and participant-generated count noise. Second, \emph{is masking useful for enforcing a fixed batch size?} We address both questions in the DP-SGD-style setting of Section~\ref{sec:distributed} with the Gaussian mechanism used as the core algorithm.

\begin{figure}[htbp]
  \centering
  \begin{subfigure}{\linewidth}\centering
    \includegraphics[width=\linewidth]{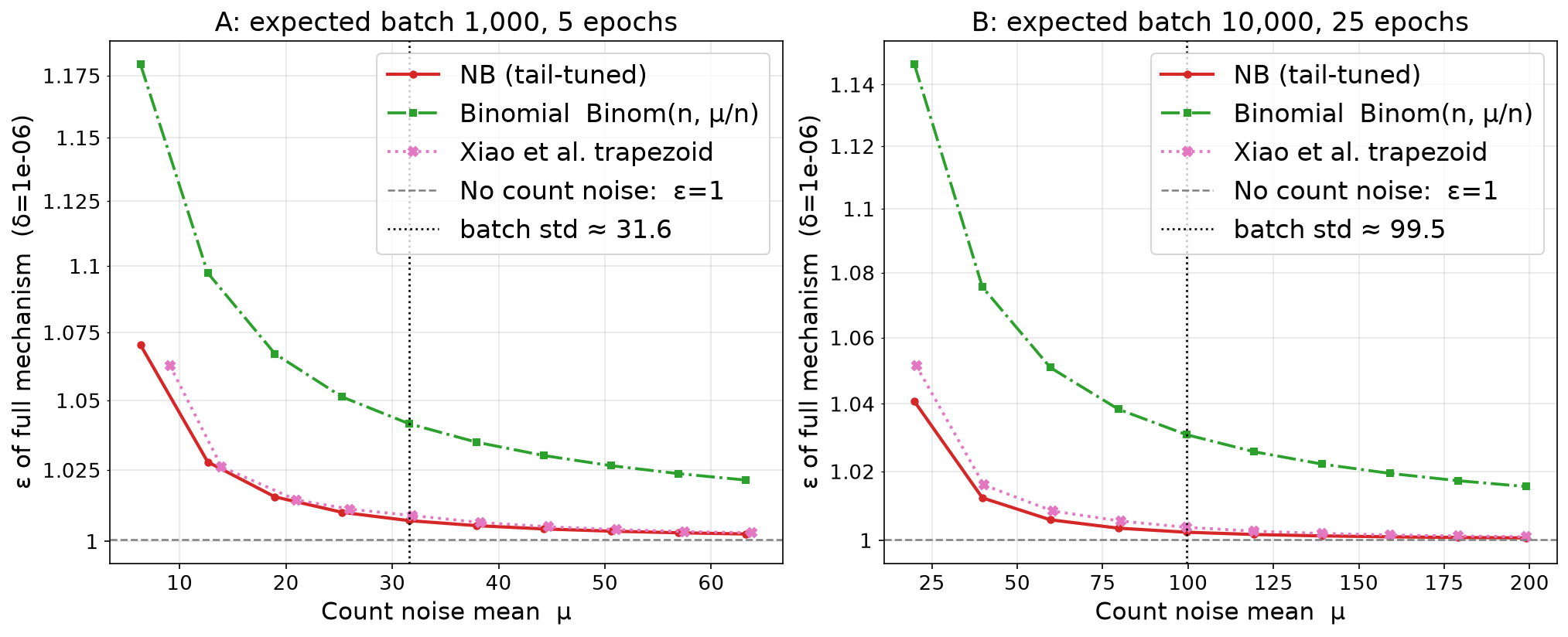}
    \caption{Poisson subsampling $\PoisMask{t,\gamma}{C}{\alg}$.}
    \label{fig:exp-overhead-pois}
  \end{subfigure}\\[0.6em]
  \begin{subfigure}{\linewidth}\centering
    \includegraphics[width=\linewidth]{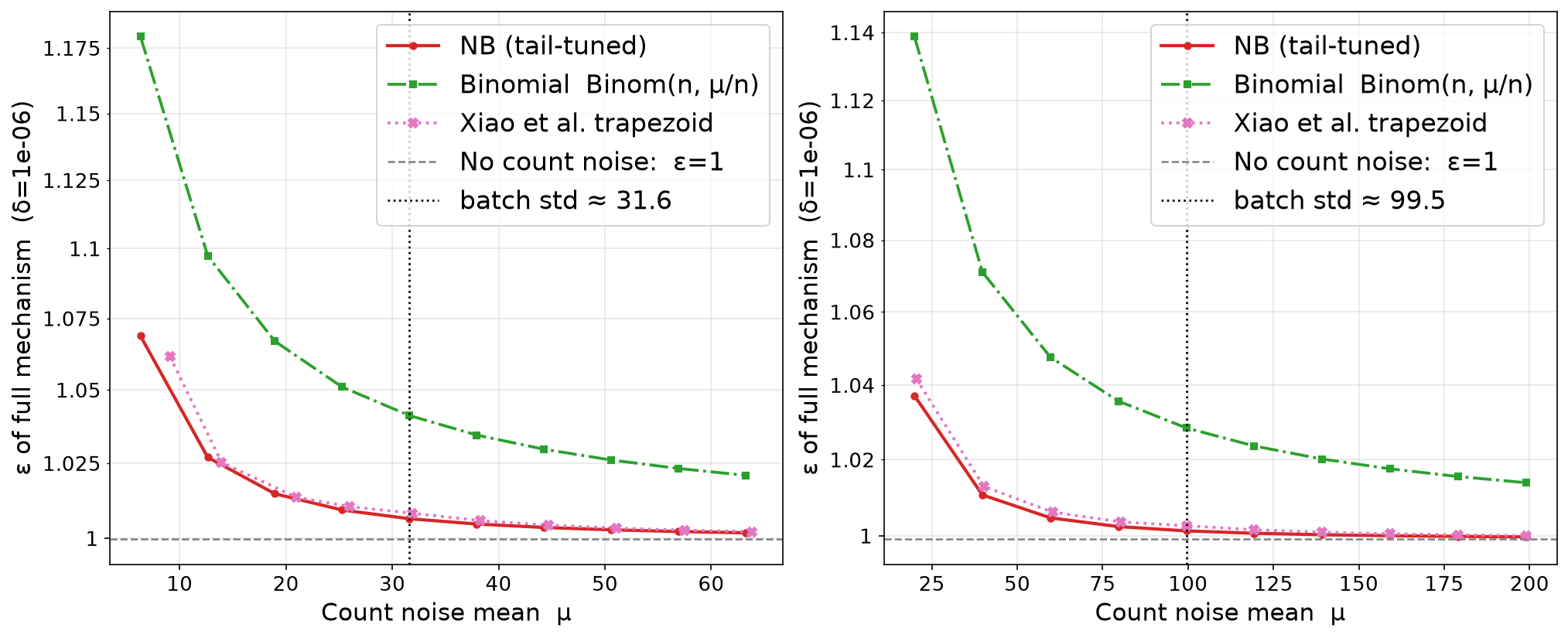}
    \caption{Random allocation $\allocMask{t,k}{C}{\alg}$.}
    \label{fig:exp-overhead-alloc}
  \end{subfigure}
  \caption{Overhead of count masking: $\eps$ at $\delta=10^{-6}$ of the full masked, composed mechanism versus count noise mean $\mu=\E[R]$, for (a)~Poisson subsampling and (b)~random allocation. The dotted vertical marks the sampler's own count standard deviation.}
  \label{fig:exp-overhead}
\end{figure}

 We compare three count-noise families: the binomial distribution $\Bin(n,q)$, with $n$ equal to the dataset size; a negative-binomial family whose parameters are tuned to minimize the resulting privacy parameter $\eps$ at the prescribed mean; and the trapezoid family of \citet{XWSD25}. We run this for both Poisson subsampling and random allocation, with one panel per setting (Figure~\ref{fig:exp-overhead}). For each prescribed mean, the tuning searches a finite grid of negative-binomial tail rates, refined around the best grid point and extended whenever the optimum would otherwise fall on a search boundary; the reported values are therefore the best found by that search rather than certified minima over the family.

We use $n=10^{6}$ users, target $\delta=10^{-6}$, and two settings of the sampling rate and number of steps: Setting~$(A)$ has rate $1/1000$ and $5{,}000$ steps (an expected batch of $1000$ and 5 epochs), and Setting~$(B)$ has rate $1/100$ and $2{,}500$ steps (an expected batch of $10{,}000$ and 25 epochs). For random allocation, the implementation accounts one independent 1-out-of-1,000 allocation per epoch and composes it five times in Setting~$(A)$, and one independent 1-out-of-100 allocation per epoch composed 25 times in Setting $(B)$. In each setting, and separately for each sampling scheme, the noise parameter $\sigma$ of the core Gaussian mechanism is calibrated so that the mechanism without dummies achieves $\eps=1$. This results in $\sigma = 0.75$ in Setting~$(A)$ and $\sigma = 2.26$ in Setting~$(B)$ for Poisson subsampling, and $\sigma = 0.74$ and $\sigma = 2.23$, respectively, for random allocation. In both cases, we compare against the baseline $\eps=1$ achieved when the counts need not be masked.%

\paragraph{The overhead of count masking}\label{sec:exp-overhead}
Figures~\ref{fig:exp-overhead-pois} (Poisson subsampling) and \ref{fig:exp-overhead-alloc} (random allocation) show the privacy cost as a function of the mean overhead $\mu=\E[R]$. The tuned negative binomial closely tracks the baseline: at an overhead equal to the standard deviation of the batch size (the dotted vertical line), it costs less than $1\%$ of extra $\eps$, and it stays close to the baseline across the whole range of $\mu$ shown. The binomial is consistently more expensive, and the trapezoid family sits close to the negative binomial. This holds for both Poisson and random-allocation sampling and in both settings; for example, under Poisson subsampling in Setting~$(A)$ the negative binomial stays within about $7\%$ of the baseline over the entire overhead range while the binomial more than doubles the overhead. Under random allocation, the negative-binomial parameters are tuned separately with the random allocation accountant, whereas the trapezoid parameters are selected by the procedure of \citet{XWSD25} (which uses their R\'enyi-DP accountant for Poisson subsampling) in both cases.

\begin{figure}[htbp]
  \centering
  \begin{subfigure}{\linewidth}\centering
    \includegraphics[width=\linewidth]{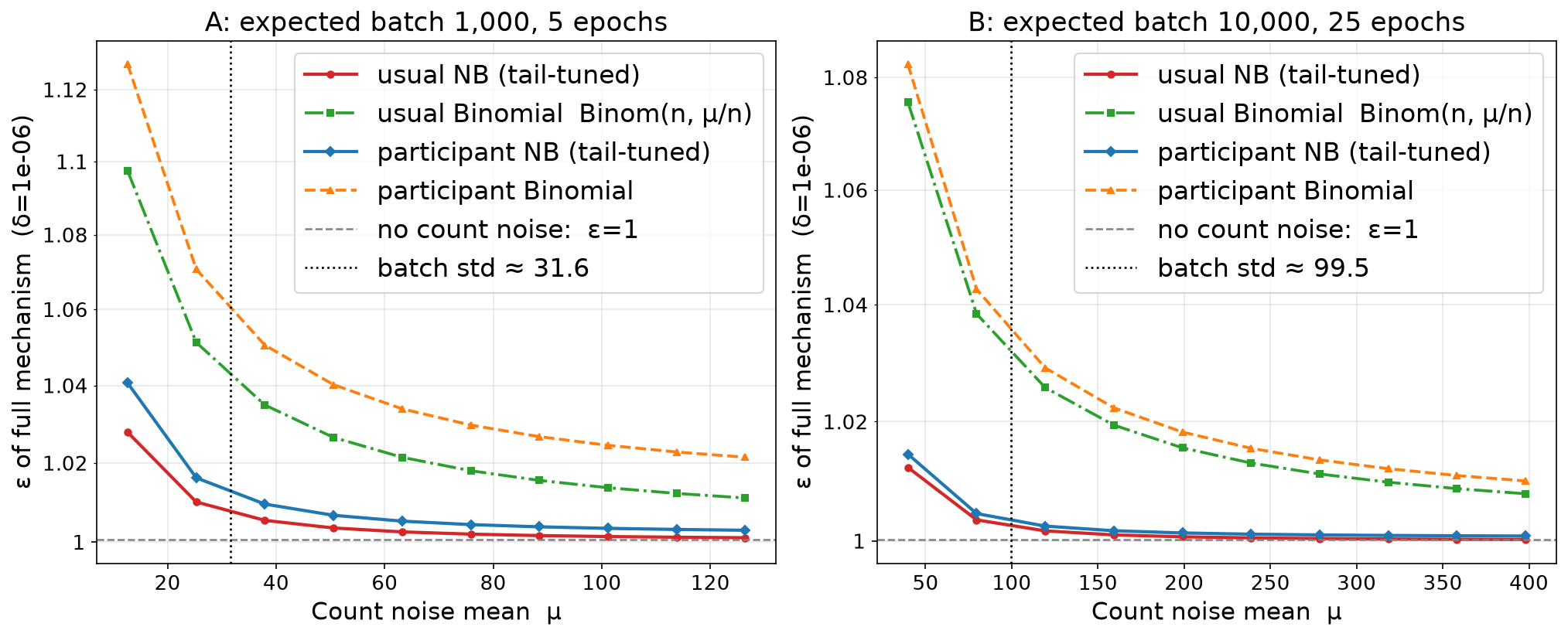}
    \caption{Poisson subsampling.}
    \label{fig:exp-participant-pois}
  \end{subfigure}\\[0.6em]
  \begin{subfigure}{\linewidth}\centering
    \includegraphics[width=\linewidth]{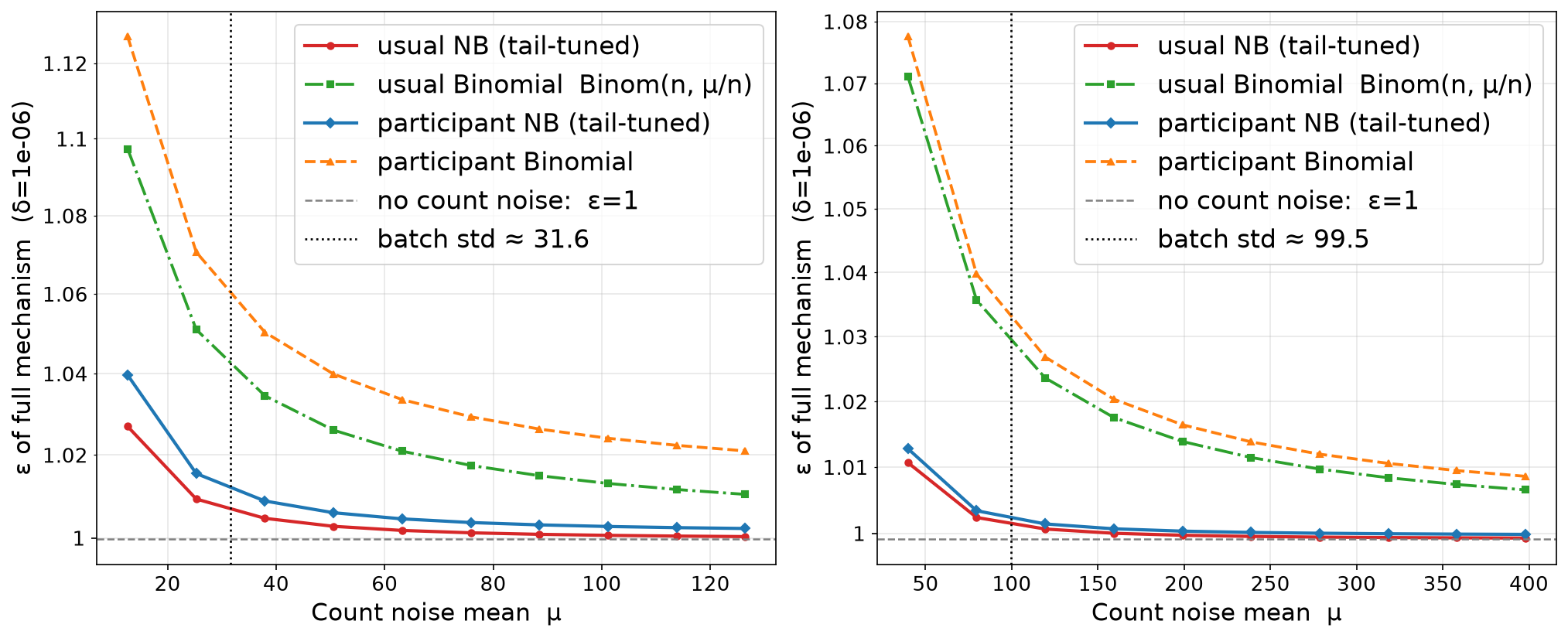}
    \caption{Random allocation.}
    \label{fig:exp-participant-alloc}
  \end{subfigure}
  \caption{Overhead of participant-generated count masking (Section~\ref{sec:participant}): $\eps$ at $\delta=10^{-6}$ of the full masked, composed mechanism versus count noise mean $\mu=\E[R]$, for (a)~Poisson subsampling and (b)~random allocation, in the same two settings and with the same $\sigma$-calibration as Figure~\ref{fig:exp-overhead}.
  The participant curves use the guaranteed minimum participation $r$ ($r=789$ in Setting~$(A)$, $r=9327$ in Setting~$(B)$) with failure budget $\beta=0.01\,\delta$.}
  \label{fig:exp-participant}
\end{figure}

\paragraph{Participant-generated dummies}\label{sec:exp-participant}
Next we compare the overhead of the participant-generated dummy count noise (as described in Section~\ref{sec:participant}) with the independent dummy count noise in the same settings. We only consider the negative binomial and the binomial noise distributions since those can be generated in a distributed manner. The accounting in this case must guarantee, with high probability, at least $r$ real participants in every batch, and must account for the removal of the deleted user's own dummies. Figure~\ref{fig:exp-participant} plots the privacy of the two noise distributions on the same axes as Figure~\ref{fig:exp-overhead}. The guaranteed number of participants $r$ is a large fraction of the mean batch,  with the failure budget set to $\beta=0.01\,\delta$ we get $r=789$ of $\E[m]=1000$ in Setting~$(A)$ and $r=9327$ of $10^{4}$ in Setting~$(B)$. In Setting~$(A)$, once the noise mean $\mu$ exceeds one standard deviation of the batch size, the privacy of the participant-generated negative binomial and binomial stays within about $0.5\%$ and $1.5\%$, respectively, of that of the corresponding independent count noise. In Setting~$(B)$ the privacy guarantees for these two types of noise generation are essentially identical across the whole range of $\mu$. Thus participant-generated dummies recover almost all of the benefit of count masking.

\paragraph{Truncation with count masking}\label{sec:exp-efficiency}
The second evaluation asks whether enforcing a hard maximum batch size $s$ is cheaper with count masking than with plain truncation/padding. A key metric for such a comparison is the \emph{batch efficiency} (Equation~\ref{eq:efficiency}): the expected fraction of a size-$s$ batch that consists of a real sampled elements (that is not dummies). Without dummy count masking, dummies are still used for padding and therefore efficiency is typically less than 1. Dummy masking adds additional dummies before truncation and therefore can reduce batch efficiency. We compare the privacy cost $\eps$ of plain truncation with that of truncation after masking with the tuned negative binomial, which gave the best trade-off above.

Formally, we define the \emph{batch efficiency} for a given batch size $s$, non-negative count noise distribution $C$ over $\NNI$ and sampling rate $\gamma$ as
\begin{equation}
  e \;=\; \E\!\left[\frac{|S|}{\max(|S|+R,\,s)}\right],
  \qquad |S|\sim\Bin(n,\gamma),\ R\sim C,
  \label{eq:efficiency}
\end{equation}
i.e.\ the pre-truncation batch (padded up to $s$) has $\max(|S|+R,s)$ items, of which $|S|$ carry a real element. Note that setting $R=0$ recovers the efficiency of plain truncation.

\begin{figure}[htbp]
  \centering
  \includegraphics[width=\linewidth]{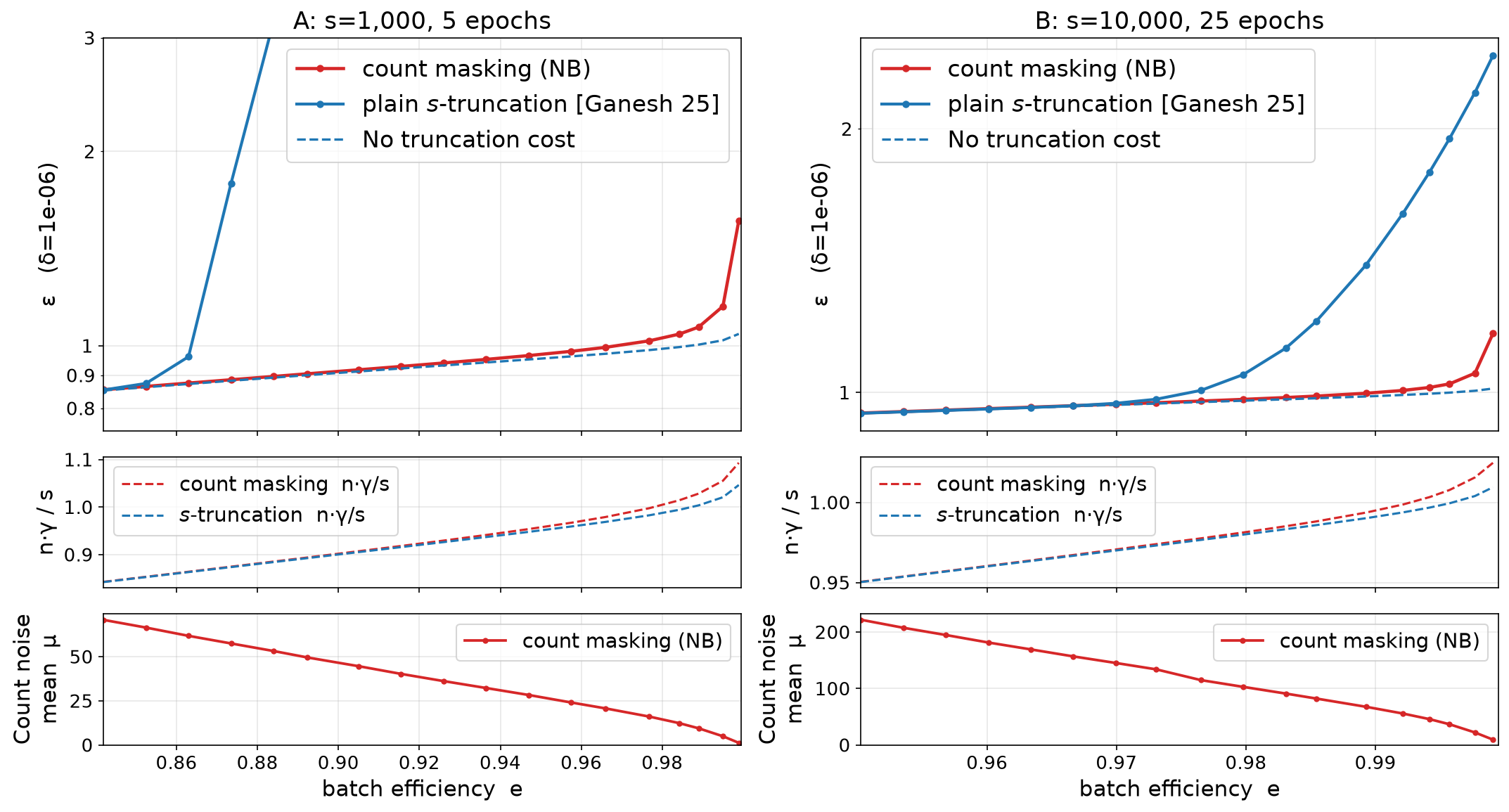}
  \caption{Count masking vs plain truncation under Poisson subsampling: $\eps$ at $\delta=10^{-6}$ (log scale) versus batch efficiency, for Setting~A ($s=1000$, left) and Setting~B ($s=10000$, right). Blue solid: plain $s$-truncation with no dummies, accounted via \citep{Ganesh25}; in Setting~A the $\eps$ axis is capped at $3$, so the upper part of this curve (rising to $\approx16$) is outside the panel. Blue dashed: basic Poisson subsampling at the same rate without the truncation cost. Red: count masking with the negative-binomial parameters and sampling rate $\gamma$ jointly searched at each efficiency (best candidate found on the grid described in the text). Lower panels: the sampling factor $n\gamma/s$ (with plain truncation's overlaid, blue dashed) and the mean count-noise magnitude $\E[R]$ resulting from the parameters chosen by the grid search.}
  \label{fig:exp-efficiency}
\end{figure}

For every value of the batch efficiency, we search for the parameters of the count masking mechanism (the sampling rate and the negative-binomial parameters) that minimize $\eps$; for plain truncation we simply select the lowest sampling rate that achieves the desired efficiency. The search is over a finite grid, refined locally around the best candidate; the sampling rate is parameterized by its relative excess over the smallest feasible rate, and the grid is extended whenever the best candidate would otherwise sit at its edge. Each reported point is additionally checked against a set of perturbed feasible candidates, none of which improves on it. We also include the privacy cost of the untruncated Poisson mechanism at the same sampling rate as a baseline for comparison. The resulting $\eps$ is plotted against batch efficiency in Figure~\ref{fig:exp-efficiency} where we also plot the mean count-noise magnitude $\E[R]$ used in the masking and the sampling factor $n\gamma/s$.

At low efficiency the two mechanisms use nearly the same sampling rate and have nearly the same privacy cost, since the batch rarely overflows (the number of real samples and dummies is below $s$ with high probability). As the efficiency increases, overflow becomes more likely and the cost of plain truncation grows rapidly, reaching $\eps\approx 15.5$ in Setting~$(A)$; the growth is especially steep in Setting~$(A)$, where the sensitivity-two component amplifies poorly at $\sigma=0.75$. Masking, in contrast, remains close to the baseline over most of the range. In both settings, a transition to oversampling (when $\gamma n > s$) occurs at higher efficiency, which is possible since dummies ensure the privacy of truncation.
Overall, we observe that at high efficiency count noise masking can significantly reduce the privacy cost of truncation.

\subsubsection*{AI Use Disclosure}
The authors formulated the problem and initiated the research of all questions addressed in this work. All parts of this paper were partially created through an iterative process in which AI (Claude 4.8/5 Opus) generated the details of proofs and wrote parts of the prose based on author-provided structure and instructions. The resulting text was also extensively edited and proofread by the authors. Code for the empirical evaluation and plotting was written by AI under the supervision and auditing checks by the authors. The authors take full responsibility for the content of this paper.

\appendix
\section{Deferred proofs for Section~\ref{sec:onesided}}\label{app:onesided}

\subsection{Proof of Lemma~\ref{lem:onesided-geometric}}\label{app:onesided-geometric}

\begin{proof}[Proof of Lemma~\ref{lem:onesided-geometric}]
Write $C=C_{b,m}$ and recall $z=1-1/b$. Splitting the normalizing sum at the mode gives
\[
    Z
      =\sum_{k=0}^{m}z^{m-k}+\sum_{k=m+1}^{\infty}z^{k-m}
      =\sum_{j=0}^{m}z^j+\sum_{j=1}^{\infty}z^j
      =\frac{1+z-z^{m+1}}{1-z},
\]
whose numerator lies in $[1,2]$; since $(1-z)^{-1}=b$, this gives $Z\in[b,2b]$ and hence
\[
    C(0)=\frac{z^m}{Z}\le\frac{z^m}{b}\le\frac{e^{-m/b}}{b},
\]
using $z=1-1/b\le e^{-1/b}$. For the mean, the expected displacement from the mode satisfies
\[
    \E_{C}[|R-m|]
      =\frac1{Z}\left(\sum_{j=1}^{m}jz^j+\sum_{j=1}^{\infty}jz^j\right)
       \le\frac{2}{Z}\sum_{j=1}^{\infty}jz^j
       =\frac{2z}{Z(1-z)^2}
       \le2b,
\]
using $Z\ge b$ and $(1-z)^{-2}=b^2$, while the displacement is upward only on the tail above the mode:
$\E_C[R]-m=\frac1Z\bigl(\sum_{j\ge1}jz^j-\sum_{j=1}^{m}jz^j\bigr)=\frac1Z\sum_{j>m}jz^j\ge0$.
Since this quantity is also at most $\E_C[|R-m|]$, we get $m\le\E_C[R]\le m+2b$, completing part~\eqref{item:onesided-shape}.

For part~\eqref{item:onesided-ratio}, $C_{+1}(0)=0$ and $C_{+1}(k)=C(k-1)$ for $k\ge1$, and the exponent $|k-1-m|-|k-m|$ equals $+1$ for $1\le k\le m$ and $-1$ for $k>m$, so $C_{+1}/C$ takes the stated values $0$, $z$ and $z^{-1}$.

Part~\eqref{item:onesided-dp} follows at once: with $e^{\eps_0}=1/z$ the ratio $C_{+1}/C$ never exceeds $e^{\eps_0}$, so $[C_{+1}-e^{\eps_0}C]_+\equiv0$; in the other direction $C/C_{+1}$ equals $+\infty$ at $k=0$, exactly $e^{\eps_0}$ on $\{1,\dots,m\}$ and $z$ beyond, so $[C-e^{\eps_0}C_{+1}]_+$ is supported on $\{0\}$, where it equals $C(0)$.

For part~\eqref{item:onesided-chi}, part~\eqref{item:onesided-ratio} gives, with $A:=\sum_{k=1}^mC(k)$ and $B:=\sum_{k=m+1}^{\infty}C(k)$,
\[
    \chi^2(C_{+1}\|C)
      =C(0)+(1-z)^2A+(z^{-1}-1)^2B
      =C(0)+\frac{A}{b^2}+\frac{B}{(b-1)^2},
\]
using $1-z=1/b$ and $z^{-1}-1=1/(b-1)$. Since $A+B\le1$ and $b^{-2}\le(b-1)^{-2}\le4b^{-2}$ for $b\ge2$, this is at most $C(0)+4/b^2$.
\end{proof}

\subsection{Proof of Proposition~\ref{prop:onesided-blackbox}}\label{app:onesided-blackbox}

We first state the effect of subsampling on hockey-stick divergences in the two-sided form that the proposition needs. For a pair $(P,Q)$ and $\gamma\in(0,1]$ write $P_\gamma:=(1-\gamma)Q+\gamma P$ as in Section~\ref{sec:onesided}; recall that $(P_\gamma,Q)$ and $(Q,P_\gamma)$ are the dominating pairs of one subsampled step (Lemma~\ref{lem:subsamPLDdom}). Both directions follow from advanced joint convexity of the hockey-stick divergence \cite[Theorem~2]{BBG18}, the second not written there verbatim; we give the two-line derivation for completeness.

\begin{lem}[Two-sided amplification by Poisson subsampling]\label{lem:onesided-amplification}
Let $\kappa\ge1$, $\kappa_\gamma:=1+\gamma(\kappa-1)$, and $\theta:=\kappa_\gamma/\kappa\le1$. Then
\[
    \HockeyStickDiv{\kappa_\gamma}{P_\gamma}{Q}=\gamma\,\HockeyStickDiv{\kappa}{P}{Q},
    \qquad
    \HockeyStickDiv{\kappa_\gamma}{Q}{P_\gamma}\le\gamma\theta\,\HockeyStickDiv{\kappa}{Q}{P}.
\]
In particular, if $(P,Q)$ is $(\eps_0,\delta_0)$-DP in both directions, then $(P_\gamma,Q)$ is $(\eps_1,\gamma\delta_0)$-DP in both directions, where $\eps_1=\log\kappa_\gamma=\log(1+\gamma(e^{\eps_0}-1))$.
\end{lem}
\begin{proof}
The first equation is immediate pointwise: $P_\gamma-\kappa_\gamma Q=\gamma(P-\kappa Q)$, so the positive parts agree after scaling by $\gamma$. For the second equation, write both arguments as mixtures with the same first component,
\[
    Q=(1-\gamma)Q+\gamma Q,
    \qquad
    P_\gamma=(1-\gamma)Q+\gamma P,
\]
and apply advanced joint convexity \cite[Theorem~2]{BBG18} to obtain the exact identity
\[
    \HockeyStickDiv{\kappa_\gamma}{Q}{P_\gamma}
      =\gamma\,\HockeyStickDiv{\kappa}{Q}{(1-\theta)Q+\theta P}.
\]
Writing $Q=(1-\theta)Q+\theta Q$ and using joint convexity of the hockey-stick divergence together with $\HockeyStickDiv{\kappa}{Q}{Q}=0$ for $\kappa\ge1$ bounds the right-hand side by $\gamma\theta\,\HockeyStickDiv{\kappa}{Q}{P}$. The final claim is the case $\kappa=e^{\eps_0}$, using $\theta\le1$ and $\HockeyStickDiv{\kappa}{\cdot}{\cdot}\le\delta_0$ in both directions.
\end{proof}

\begin{proof}[Proof of Proposition~\ref{prop:onesided-blackbox}]
Let $\eps_0:=1/(8\sigmastar)$ and $\delta_0:=\delta/(2\gamma t)$; by the asymptotic conditions~\eqref{eq:onesided-asymptotics} we may assume $\eps_0\le\log2$, so that $b:=1/(1-e^{-\eps_0})$ satisfies $2\le b\le 1+1/\eps_0\le2/\eps_0=16\sigmastar$. Let $C=C_{b,m}$ with this $b$ (so that $z=e^{-\eps_0}$) and the first calibration of the center $m$ (eq.~\eqref{eq:onesided-calibrations}), which gives $C(0)\le\delta_0$. By Lemma~\ref{lem:onesided-geometric}\eqref{item:onesided-dp}, $C$ is then $(\eps_0,\delta_0)$-DP in both directions, so by Lemma~\ref{lem:onesided-amplification} one subsampled release is $(\eps_1,\gamma\delta_0)$-DP in both directions, where
\[
    \eps_1=\log\bigl(1+\gamma(e^{\eps_0}-1)\bigr)\le\gamma(e^{\eps_0}-1)\le2\gamma\eps_0=\frac{\gamma}{4\sigmastar},
\]
using $e^x-1\le2x$ for $x\in[0,1]$. Advanced composition \citep{DRV10,DR14} over the $t$ releases, with an additional failure probability $\delta/2$, then gives total privacy parameter at most
\[
    \eps_1\sqrt{2t\log(2/\delta)}+t\eps_1(e^{\eps_1}-1)
    \le \frac{\gamma}{4\sigmastar}\sqrt{4t\logdelta }+\frac{2t\gamma^2}{16{\sigmastar}^2}
    = \frac{\eps}{2\sqrt2}+\frac{\eps^2}{16\logdelta }
    \le\eps,
\]
using $\log(2/\delta)\le2\logdelta $ (valid as $\logdelta \ge1$), $\eps_1\le1$ and $\eps\le \logdelta $, while the total failure probability is $t\gamma\delta_0+\delta/2=\delta$. Since $(P_\gamma^{\otimes t},Q^{\otimes t})$ is itself the $t$-fold (non-adaptive) composition of the pair $(P_\gamma,Q)$, advanced composition applies to it directly in both orientations, and hence $\delta_C^{\gamma,t}(\eps)\le\delta$. Finally, by the first calibration of the center (eq.~\eqref{eq:onesided-calibrations}), Lemma~\ref{lem:onesided-geometric}\eqref{item:onesided-shape}, and $b\le16\sigmastar$,
\[
    \E[R]\le b\left[\log\frac1{b\delta_0}\right]_++2b+1
      \le b\log\frac1{\delta_0}+2b+1
      = b\log\frac{2\gamma t}{\delta} + 2b +1=O\!\left(\sigmastar\log\frac{\gamma t}\delta\right),
\]
since $\log\frac{\gamma t}{\delta}\ge1$ and $\gamma t\ge1$.
\end{proof}

\subsection{Proof of Theorem~\ref{thm:onesided-gaussian}}\label{app:onesided-gaussian}

Our proof relies on the following form of the $\eps$ of GDP up to lower-order terms.
\begin{lem}[$\eps$ of GDP up to lower-order terms]\label{lem:onesided-gdp-asymp}
As $\delta\to0$ and $V\ge\delta^{o(1)}$,
\[
    \eps_{\mathrm G}(V,\delta)=\frac V2+(1+o(1))\sqrt{2V\log(1/\delta)} .
\]
\end{lem}
\begin{proof}
Let $L:=\log(1/\delta)$. The upper bound is the bound on $\eps_{\mathrm G}$ in eq.~\eqref{eq:onesided-gdp-tail}. For the lower bound let $\eta:=\min\{\sqrt V,1\}$ and $z:=\sqrt{2(L-\log(8L/\eta))}$, which satisfies $z=(1-o(1))\sqrt{2L}$ and $z\ge1$ in this regime, and let $\eps':=V/2+\sqrt V z-\eta$. The privacy loss $\Lambda$ of $\sqrt V$-GDP has distribution $N(V/2,V)$ \cite[Proposition~3]{Steinke22}, and $\HockeyStickDiv{e^{\eps'}}{N(\sqrt V,1)}{N(0,1)}=\E\bigl[[1-e^{\eps'-\Lambda}]_+\bigr]$, where $[1-e^{\eps'-\Lambda}]_+\ge1-e^{-\eta}\ge\eta/2$ whenever $\Lambda\ge\eps'+\eta$. Denoting by $\varphi$ the standard normal density and by $\bar\Phi$ its tail, Gordon's inequality $\bar\Phi(z)\ge\varphi(z)\,z/(1+z^2)\ge\varphi(z)/(2z)$ \citep{Gordon41} therefore gives
\[
    \HockeyStickDiv{e^{\eps'}}{N(\sqrt V,1)}{N(0,1)}
      \ \ge\ \frac\eta2\,\bar\Phi(z)
      \ \ge\ \frac{\eta\,\varphi(z)}{4z}
      \ =\ \frac{2L\,\delta}{\sqrt{2\pi}\,z}
      \ \ge\ \sqrt{\frac L\pi}\;\delta
      \ >\ \delta,
\]
where we used that $z\le\sqrt{2L}$ and $L>\pi$. Thus $\eps_{\mathrm G}(V,\delta)>\eps'\ge V/2+\sqrt V(z-1)=V/2+(1-o(1))\sqrt{2VL}$.
\end{proof}

\begin{proof}[Proof of Theorem~\ref{thm:onesided-gaussian}]
Since $\sigma\ge1$ and $\alpha\le1$ we have $b\ge\sqrt{80}>2$, so Lemma~\ref{lem:onesided-geometric} applies and the second calibration of the center $m$ (eq.~\eqref{eq:onesided-calibrations}) gives
\begin{equation}\label{eq:onesided-alpha-budget}
    \chi_c^2\ \le\ \frac{e^{-m/b}}{b}+\frac{4}{b^2}\le\frac5{b^2}\ \le\ \frac{\alpha}{16\sigma^2}\ \le\ \frac{\alpha}{8}\left(1-e^{-1/\sigma^2}\right)
    \ =\ \frac{\alpha}{8}\cdot\frac{\chi_s^2}{1+\chi_s^2},
\end{equation}
using $1-e^{-u}\ge u/2$ for $u\in(0,1]$ with $u=1/\sigma^2$, and the $\chi^2$ divergence of the Gaussian pair (eq.~\eqref{eq:onesided-gauss-chi}) for the last step.

By the factorization $1+\chi^2(P\|Q)=(1+\chi_c^2)(1+\chi_s^2)$ and \eqref{eq:onesided-alpha-budget},
\[
    v:=\chi^2(P\|Q)=\chi_s^2+\chi_c^2(1+\chi_s^2)\ \le\ \Bigl(1+\frac\alpha8\Bigr)\chi_s^2 .
\]
In both schemes the variance in the CLT condition is proportional to the $\chi^2$ divergence of the pair. Therefore the variance $V$ in the CLT condition for $(P,Q)$ satisfies $V\le(1+\alpha/8)V_0$. By the CLT condition for $(P,Q)$ and Corollaries~\ref{cor:poissonamp} and~\ref{cor:RAmain}, the masked release is $\bigl(\eps_{\mathrm G}(V,\delta),\delta\bigr)$-DP in both directions. Using the bound on $\eps_{\mathrm G}$ (eq.~\eqref{eq:onesided-gdp-tail}) and then Lemma~\ref{lem:onesided-gdp-asymp} for $V_0$, we obtain that
\[
    \eps_{\mathrm G}(V,\delta)
      \ \le\ \frac V2+\sqrt{2V\logdelta}
      \ \le\ \Bigl(1+\frac\alpha8\Bigr)\Bigl(\frac{V_0}{2}+\sqrt{2V_0\logdelta}\Bigr)
      \ =\ (1+o(1))\Bigl(1+\frac\alpha8\Bigr)\eps .
\]
For a fixed $\alpha$ this is at most $(1+\alpha)\eps$ for all sufficiently large parameters. The mean bound is Lemma~\ref{lem:onesided-geometric}\eqref{item:onesided-shape} with $m\le b\log b+1$.

Finally, $\eps\ge V_0/2$ by Lemma~\ref{lem:onesided-gdp-asymp} and $\eps\le\logdelta$ by the asymptotic conditions~\eqref{eq:onesided-asymptotics}. Thus $V_0\le2\logdelta$, that is, $V_0/2\le\sqrt{2V_0\logdelta}/2$, and Lemma~\ref{lem:onesided-gdp-asymp} gives $(1-o(1))\sqrt{2V_0\logdelta}\le\eps\le\frac32\sqrt{2V_0\logdelta}$. Therefore $\sigmastar=\gamma\sqrt{2t\logdelta}/\eps=\Theta(\gamma\sqrt{t/V_0})$. For Poisson subsampling $\gamma\sqrt{t/V_0}=\sqrt{1-\gamma}/\chi_s$ with $\sqrt{1-\gamma}\ge1/\sqrt2$, and for random allocation $\gamma\sqrt{t/V_0}=1/\chi_s$ since $\gamma=1/t$. In addition, $1/\sigma^2\le\chi_s^2=e^{1/\sigma^2}-1\le2/\sigma^2$ for $\sigma\ge1$, using $x\le e^x-1\le2x$ for $x\in[0,1]$. Therefore $\sigma=\Theta(\sigmastar)$. Since $b\le\sigma\sqrt{80/\alpha}+1$, the mean bound is then also $O(\frac{\sigmastar}{\sqrt\alpha}\log\frac{\sigmastar}{\sqrt\alpha})$.
\end{proof}

\printbibliography
\end{document}